\documentclass[10pt,twocolumn,letterpaper]{article}

\usepackage[pagenumbers]{cvpr}

\input{preamble}

\definecolor{cvprblue}{rgb}{0.21,0.49,0.74}
\usepackage[pagebackref,breaklinks,colorlinks,allcolors=cvprblue]{hyperref}

\hypersetup{pdftitle={RadAG: Reward-Adversarial Attention Guidance for Efficient Inference-Time Alignment of Diffusion Models},pdfauthor={Hyeongchan Jeon, Kwanyoung Kim}}

\def\confName{CVPR}
\def\confYear{2027}

\title{RadAG: Reward-Adversarial Attention Guidance for Efficient Inference-Time Alignment of Diffusion Models}

\author{Hyeongchan Jeon \qquad Kwanyoung Kim$^{\dagger}$\\
Department of AI, Gwangju Institute of Science and Technology (GIST)\\
{\tt\small rabo8497@gm.gist.ac.kr \qquad k0.kim@gist.ac.kr}}

\makeatletter
\apptocmd{\@maketitle}{\begin{center}
\begin{minipage}{\textwidth}
\centering
\includegraphics[width=\linewidth]{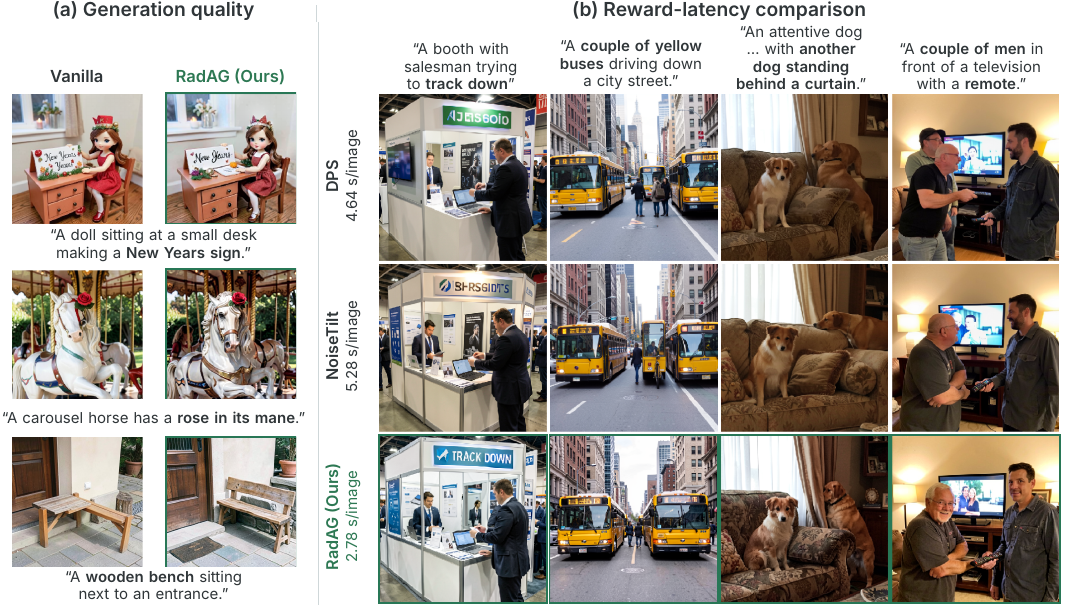}
\captionsetup{hypcap=false}
\captionof{figure}{\rev{\textbf{RadAG aligns a diffusion model with a reward at low cost.}
(a) Samples of the base model (Vanilla) and RadAG from the same prompts and seeds. (b) Samples of DPS~\cite{dps}, NoiseTilt~\cite{noisetilt}, and RadAG with their time per image; across our settings, RadAG takes 30--65\% less time than gradient-based and candidate-based baselines (Sec.~\ref{sec:main}).}}
\label{fig:teaser}
\end{minipage}
\end{center}
\vspace{0.5em}
}{}{\PackageError{paper}{Could not insert teaser}{Check the title macro.}}
\makeatother

\begin{document}
\maketitle
{\renewcommand{\thefootnote}{\fnsymbol{footnote}}\footnotetext[2]{Corresponding author.}}
\begin{figure}[t]
\centering
\includegraphics{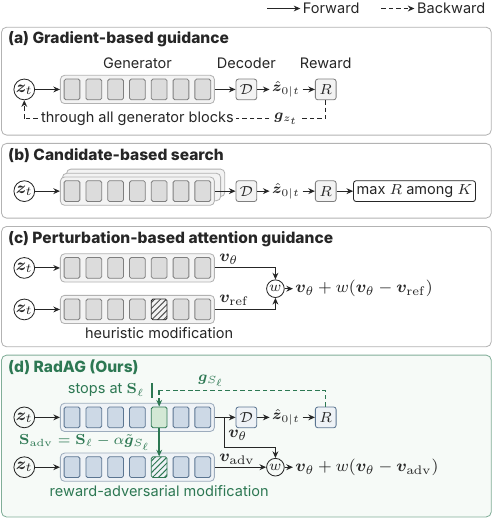}
\caption{\rev{\textbf{Comparison of RadAG with existing methods.} (a) Gradient-based guidance backpropagates the reward through the entire generator, (b) candidate-based search selects among $K$ samples, and (c) perturbation-based attention guidance modifies attention by a heuristic rule. (d) RadAG stops the reward gradient at the attention logits of block $\ell$ and moves them against it.}}
\label{fig:overview}
\end{figure}

\begin{abstract}
\rev{Diffusion and flow-based models have achieved remarkable success in generation, yet their samples often remain suboptimal in fidelity and text alignment.}
\rev{Two lines of inference-time methods address this: extrapolation guidance, such as classifier-free guidance (CFG) and its variants, and reward guidance, which aligns samples with reward models of human preference.}
\rev{Extrapolation guidance needs only forward passes, but it builds its reference by heuristic rules unrelated to the reward and therefore cannot steer sampling toward a given reward.}
\rev{Gradient-based reward guidance steers sampling directly toward a higher reward, but backpropagates through the entire generator at every step.}
\rev{This trade-off raises the question of \emph{whether extrapolation guidance can steer sampling toward a higher reward without a full backward pass}.}
\rev{To answer this, we introduce Reward-Adversarial Attention Guidance (RadAG), which constructs the reference by perturbing the attention logits of a single block against their reward gradient and extrapolates away from it.}
\rev{We prove that this update increases the reward of the clean estimate to first order, while its backward pass stops at the perturbed block instead of traversing the entire generator.}
\rev{On FLUX.2-klein-4B and 9B with PickScore and aesthetic rewards, RadAG achieves the highest target reward in every setting with 30--65\% less time than gradient-based and candidate-based baselines.}
\end{abstract}

\section{Introduction}
\label{sec:intro}

\rev{Diffusion and flow-based models have achieved remarkable success in image generation~\cite{ddpm,flowmatching,sd3} and are rapidly extending to video~\cite{wan,hunyuanvideo}.}
\rev{Without guidance, however, their samples are often suboptimal, losing detail or missing parts of the prompt.}

\rev{Inference-time guidance addresses this by modifying the prediction during sampling, and classifier-free guidance (CFG)~\cite{cfg}, which extrapolates the conditional prediction away from an unconditional one, is the dominant approach.}
\rev{CFG has inspired a family of extrapolation guidance methods that replace the unconditional reference with other degraded predictions, obtained from a less-trained version of the model~\cite{autoguidance} or from attention-based perturbations of the model~\cite{sag,pag,seg,asag}.}
\rev{These methods improve samples using forward passes only, but their references follow heuristic rules unrelated to the reward, so they cannot steer sampling toward a given reward.}

\rev{Reward guidance offers an alternative that steers sampling toward a higher reward directly, using reward models of human preference~\cite{pickscore,imagereward,laionaes}.}
\rev{Unlike fine-tuning~\cite{ddpo,draft,adjointmatching}, which requires training for each reward, inference-time reward guidance keeps the generator frozen.}
\rev{Besides candidate-based search~\cite{svdd,das,dts}, whose cost grows with the number of candidates, the most direct approach is gradient-based guidance~\cite{dps,freedom,noisetilt}, which evaluates the reward on the estimated clean sample and differentiates it with respect to the noisy sample.}
\rev{Because the clean estimate is a function of the generator output, this gradient requires backpropagation through the reward model, the decoder, and the entire generator at every step (Fig.~\ref{fig:overview}).}

\begin{table}[t]
\centering
\caption{\rev{\textbf{Comparison of inference-time guidance methods.}}}
\label{tab:intro_comparison}
\begingroup
\footnotesize
\setlength{\tabcolsep}{3pt}
\renewcommand{\arraystretch}{1.12}
\newcommand{\introYes}{\enabled}
\newcommand{\introNo}{\disabled}
\begin{tabularx}{\linewidth}{@{}Xcccc@{}}
\toprule
Method & \shortstack{\scriptsize No extra\\\scriptsize training} & \shortstack{\scriptsize No candidate\\\scriptsize branching} & \shortstack{\scriptsize Reward-\\\scriptsize guided} & \shortstack{\scriptsize No full\\\scriptsize backward}\\
\cmidrule(r){1-1}\cmidrule(l){2-5}
CFG~\cite{cfg} & \introNo & \introYes & \introNo & \introYes\\
DPS~\cite{dps}, NoiseTilt~\cite{noisetilt} & \introYes & \introYes & \introYes & \introNo\\
SVDD-PM~\cite{svdd} & \introYes & \introNo & \introYes & \introYes\\
PAG~\cite{pag}, SEG~\cite{seg} & \introYes & \introYes & \introNo & \introYes\\
\cmidrule(r){1-1}\cmidrule(l){2-5}
\rowcolor{ourshighlight}
\textbf{RadAG (Ours)} & \introYes & \introYes & \introYes & \introYes\\
\bottomrule
\end{tabularx}
\endgroup
\end{table}

\rev{Extrapolation guidance is thus efficient but cannot steer toward a reward, whereas gradient-based guidance steers toward a reward at a high cost (Table~\ref{tab:intro_comparison}).}
\rev{This trade-off raises the question: \emph{Can extrapolation guidance steer sampling toward a higher reward without a full backward pass?}}

\rev{To answer this question, we first examine what the reference of extrapolation guidance moves away from.}
\rev{Visualizing attention, we find that existing references perturb it by rules unrelated to the image content, whereas the reward is most sensitive to the attention logits on the objects of each sample, where the base attention also concentrates.}

\begin{figure}[t]
\centering
\includegraphics{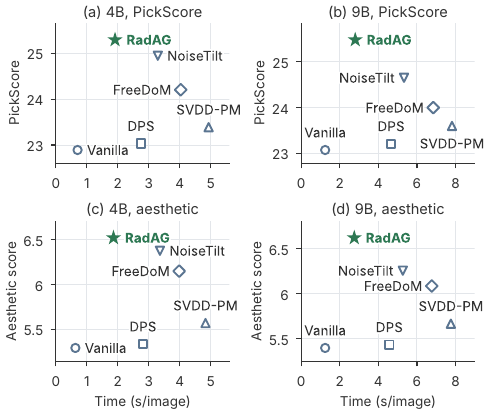}
\caption{\rev{\textbf{Reward--time trade-off of reward-based methods.} Target reward versus generation time for Table~\ref{tab:main}, with batch size 4 (4B) and 2 (9B) on one NVIDIA RTX A6000; upper left is better.}}
\label{fig:reward_time}
\end{figure}

\rev{Building on this observation, we propose Reward-Adversarial Attention Guidance (RadAG), which perturbs these logits in the direction that lowers the reward and extrapolates away from the resulting prediction.}
\rev{We prove that this update increases the reward of the clean estimate to first order, and since it keeps the form of CFG, existing refinements of CFG can also be applied.}
\rev{To compute the reward gradient efficiently, we further introduce a partial backward pass that stops at the perturbed block instead of traversing the entire generator, together with forward reuse that computes the reference from the cached computation before this block.}
\rev{We summarize our contributions as follows:}
\begin{itemize}
\item \rev{To steer extrapolation guidance toward a reward, we propose RadAG, which perturbs attention logits against their reward gradient to construct the reference, and prove its first-order reward ascent.}
\item \rev{RadAG stops the generator backward pass at the perturbed block and reuses the forward computation before it, which makes the generator backward pass more than 6$\times$ faster than that of DPS.}
\item \rev{Across model scales and rewards, RadAG achieves a better reward--time trade-off than existing inference-time guidance (Fig.~\ref{fig:reward_time}). Combined with existing refinements of CFG, RadAG+ raises the target reward further at no additional backward cost.}
\end{itemize}

\section{Background}
\label{sec:background}

\subsection{Diffusion and Flow-Based Generative Models}
\label{sec:flow}
\rev{Diffusion and flow-based generative models~\cite{ddpm,flowmatching,rectifiedflow} generate a clean latent $\boldsymbol{z}_0$ from Gaussian noise $\boldsymbol{\epsilon}\sim\mathcal{N}(\mathbf{0},\mathbf{I})$ by reversing a noising process.}
\rev{With the linear interpolant $\boldsymbol{z}_t=(1-t)\boldsymbol{z}_0+t\boldsymbol{\epsilon}$ for $t\in[0,1]$, flow matching~\cite{flowmatching,rectifiedflow} trains a velocity field $\boldsymbol{v}_\theta$ conditioned on $c$ by}
\begin{equation}
\revcolor \min_\theta\,\mathbb{E}_{t,\boldsymbol{z}_0,\boldsymbol{\epsilon}}\big\|(\boldsymbol{\epsilon}-\boldsymbol{z}_0)-\boldsymbol{v}_\theta(\boldsymbol{z}_t,t,c)\big\|_2^2.
\label{eq:noisy}
\end{equation}
\rev{Since the regression target is $\boldsymbol{\epsilon}-\boldsymbol{z}_0$, the prediction also gives a clean estimate,}
\begin{equation}
\revcolor \hat{\boldsymbol{z}}_{0|t}=\boldsymbol{z}_t-t\,\boldsymbol{v}_\theta(\boldsymbol{z}_t,t,c),
\label{eq:clean_estimate}
\end{equation}
\rev{the flow counterpart of Tweedie's formula for the posterior mean $\mathbb{E}[\boldsymbol{z}_0\mid\boldsymbol{z}_t,c]$~\cite{efron2011tweedie,noise2score}.}
\rev{Sampling proceeds from $t=1$ to $t=0$, and for $t'<t$ an Euler step reads}
\begin{equation}
\revcolor \boldsymbol{z}_{t'}=\boldsymbol{z}_t+(t'-t)\,\boldsymbol{v}_\theta(\boldsymbol{z}_t,t,c).
\label{eq:euler}
\end{equation}
\rev{Guidance replaces $\boldsymbol{v}_\theta$ in Eq.~\eqref{eq:euler} with a modified prediction.}
\rev{A VAE decoder $\mathcal{D}$~\cite{ldm} maps the clean estimate to the image $\hat{\boldsymbol{x}}_{0|t}=\mathcal{D}(\hat{\boldsymbol{z}}_{0|t})$ for reward evaluation.}

\subsection{Classifier-Free and Perturbation-Based Attention Guidance}
\label{sec:perturbation}
\rev{Classifier-free guidance (CFG)~\cite{cfg} trains the generator with condition dropout so that it also predicts $\boldsymbol{v}_\theta(\boldsymbol{z}_t,t,\varnothing)$ under the null condition $\varnothing$.}
\rev{Abbreviating $\boldsymbol{v}_\theta(\boldsymbol{z}_t,t,c)$ as $\boldsymbol{v}_\theta$ and taking $\boldsymbol{v}_{\mathrm{ref}}=\boldsymbol{v}_\theta(\boldsymbol{z}_t,t,\varnothing)$, CFG replaces the Euler step of Eq.~\eqref{eq:euler} with}
\begin{equation}
\revcolor \boldsymbol{z}_{t'}=\boldsymbol{z}_t+(t'-t)\big[\boldsymbol{v}_\theta+w\big(\boldsymbol{v}_\theta-\boldsymbol{v}_{\mathrm{ref}}\big)\big],
\label{eq:extrapolation}
\end{equation}
\rev{where $w\geq0$ is the guidance scale.}
\rev{Since $\boldsymbol{v}_\theta-\boldsymbol{v}_{\mathrm{ref}}$ is proportional to the score of $p(c\mid\boldsymbol{z}_t)$, CFG tilts the conditional distribution toward a higher implicit reward $\log p(c\mid\boldsymbol{z}_t)$~\cite{cfg}.}
\rev{We refer to updates of the form of Eq.~\eqref{eq:extrapolation}, with any reference prediction $\boldsymbol{v}_{\mathrm{ref}}$, as extrapolation guidance.}
\rev{Other methods change the reference. For example, autoguidance~\cite{autoguidance} takes $\boldsymbol{v}_{\mathrm{ref}}$ from a less-trained version of the generator, and perturbation-based attention guidance takes it from the pretrained generator with modified attention.}
\rev{At block $\ell$, attention computes}
\begin{equation}
\revcolor \mathrm{Attn}_\ell=\texttt{Softmax}(\mathbf{S}_\ell)\,\mathbf{V}_\ell,\qquad \mathbf{S}_\ell=\mathbf{Q}_\ell\mathbf{K}_\ell^\top/\sqrt{d},
\label{eq:attention}
\end{equation}
\rev{where $\mathrm{Attn}_\ell$ is the attention output, $\mathbf{S}_\ell$ are the attention logits, $\mathbf{A}_\ell=\texttt{Softmax}(\mathbf{S}_\ell)$ is the attention map over keys, and $d$ is the key dimension per head.}
\rev{SAG~\cite{sag} blurs the input regions selected by $\mathbf{A}_\ell$, PAG~\cite{pag} replaces $\mathbf{A}_\ell$ with the identity, SEG~\cite{seg} smooths $\mathbf{A}_\ell$ by blurring the queries, and ASAG~\cite{asag} recomputes $\mathbf{A}_\ell$ adversarially with Sinkhorn iterations.}
\rev{These heuristic rules do not depend on any reward, so they cannot steer sampling toward a given reward even when one is available.}

\subsection{Gradient-Based Guidance}
\label{sec:gradient}
\rev{Let $R(\boldsymbol{x},c)$ be a differentiable scalar reward, such as a model of human preference, with larger values indicating better alignment.}
\rev{Gradient-based guidance evaluates the reward on the decoded clean estimate and differentiates it with respect to the noisy sample,}
\begin{equation}
\revcolor \boldsymbol{g}_{z_t} = \nabla_{\boldsymbol{z}_t}R\big(\mathcal{D}(\hat{\boldsymbol{z}}_{0|t}),c\big),
\label{eq:latent_grad}
\end{equation}
\rev{and adds it to the Euler step of Eq.~\eqref{eq:euler}:}
\begin{equation}
\revcolor \boldsymbol{z}_{t'}=\boldsymbol{z}_t+(t'-t)\,\boldsymbol{v}_\theta+\lambda_t\boldsymbol{g}_{z_t},
\label{eq:reward_update}
\end{equation}
\rev{where $\lambda_t\geq0$ is the correction strength, including any schedule-dependent scaling.}
\rev{This form follows diffusion posterior sampling (DPS)~\cite{dps}, which uses a measurement log-likelihood in place of the reward, and FreeDoM~\cite{freedom}, which repeats the guided step within a range of noise levels; NoiseTilt~\cite{noisetilt} instead injects a whitened reward gradient through the stochastic noise term.}
\rev{Compared with Eq.~\eqref{eq:extrapolation}, the correction $\lambda_t\boldsymbol{g}_{z_t}$ steers toward the reward directly, but because $\hat{\boldsymbol{z}}_{0|t}$ depends on the generator prediction, computing it requires backpropagation through the reward model, the decoder, and all generator blocks at every step (Fig.~\ref{fig:overview}).}
\rev{LAGS~\cite{lags} and QGF~\cite{qgf} avoid this backward pass by approximating the Jacobian of the clean estimate with a finite difference or the identity, whereas RadAG takes the exact gradient at the attention logits of a late block (App.~\ref{app:related}).}

\section{Motivation}
\label{sec:motivation}
\rev{Extrapolation guidance moves the prediction away from a reference, so what it improves depends on how the reference differs from the base prediction.}
\rev{For perturbation-based attention guidance, this difference lies in the attention of one block.}
\rev{We therefore examine what existing references change in the attention, and which attention a reference would have to change to raise a given reward.}

\begin{figure}[t]
\centering
\includegraphics[width=\linewidth]{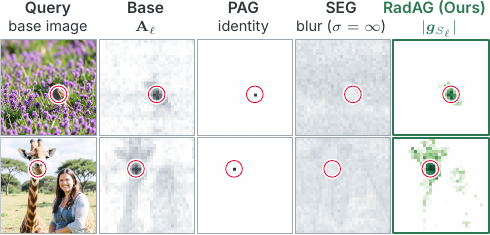}
\caption{\rev{\textbf{Attention of the references and reward sensitivity.}
Attention of an image query (circle) over the image keys in the base model, PAG, and SEG, and its reward sensitivity $|\boldsymbol{g}_{S_\ell}|$.}}
\label{fig:motivation}
\vspace{6pt}
\centering
\includegraphics{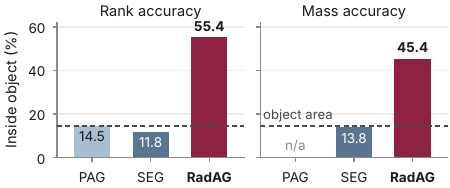}
\caption{\rev{\textbf{Concentration of the maps on the queried object.} Rank and mass accuracy of the PAG and SEG attention maps and of the reward sensitivity of RadAG against the mask of the queried object over 100 COCO prompts; the dashed line is the object area.}}
\label{fig:sensitivity}
\end{figure}

\paragraph{Content-agnostic references.}
\rev{Fig.~\ref{fig:motivation} shows, for an image query (circle) in each example, its attention over the image keys in the base model and in the references of PAG and SEG, which act on this image-to-image part of the joint attention.}
\rev{In the base model, the query on the bird attends to the bird, and the query on the giraffe attends to the giraffe.}
\rev{PAG replaces this row with the identity, so the query attends only to itself, and SEG spreads it almost uniformly over the image.}
\rev{Both apply the same heuristic to every query, prompt, and reward, so the reference, and with it the direction of improvement, is unrelated to the target reward.}

\paragraph{Reward sensitivity.}
\rev{To find which attention a reference should change instead, we measure how strongly the reward depends on each attention logit and define the reward sensitivity as the gradient of the reward of the clean estimate in Eq.~\eqref{eq:clean_estimate} with respect to the attention logits of block $\ell$,}
\begin{equation}
\revcolor \boldsymbol{g}_{S_\ell}=\nabla_{\mathbf{S}_\ell}R\big(\mathcal{D}(\hat{\boldsymbol{z}}_{0|t}),c\big).
\label{eq:logit_grad}
\end{equation}
\rev{The last column of Fig.~\ref{fig:motivation} shows its magnitude for the same query: unlike the references of PAG and SEG, it concentrates on the queried object.}

\paragraph{Concentration on objects.}
\rev{Fig.~\ref{fig:sensitivity} checks this over 100 COCO prompts.}
\rev{For each prompt, we segment the object under the query with Mask2Former~\cite{mask2former} and compare its mask with the map of the query by two standard measures for heatmaps~\cite{clevrxai}: rank accuracy, the share of the $K$ largest entries of the map that fall inside the mask, where $K$ is the mask size, and mass accuracy, the share of the sum of the map inside the mask (App.~\ref{app:concentration}).}
\rev{A map that does not prefer the object scores the object area in both, 15\% on average.}
\rev{The reward sensitivity scores 55\% and 45\%, three to four times this area, whereas the PAG and SEG maps stay at or below it, so they change the attention on the object no more than a uniform map would.}

\rev{Therefore, the reward of a sample depends mainly on the attention to its objects, which heuristic references do not target.}
\rev{If the reference instead changes exactly these logits in the direction that lowers the reward, extrapolating away from it moves the sample where the reward is decided and should raise the reward.}

\section{Method}
\label{sec:method}
\rev{Following Sec.~\ref{sec:motivation}, we construct a reference that lowers the reward through the attention logits and prove that extrapolating away from it raises the reward (Sec.~\ref{sec:perturb}), compute this reference with a partial backward pass and forward reuse (Sec.~\ref{sec:efficient}), and extend RadAG with refinements of extrapolation guidance (Sec.~\ref{sec:combine}).}

\subsection{Reward-Adversarial Attention Guidance}
\label{sec:perturb}
\paragraph{Reward-adversarial reference.}
\rev{Sec.~\ref{sec:motivation} showed that the reward depends mainly on the attention logits with a large reward sensitivity $\boldsymbol{g}_{S_\ell}$ in Eq.~\eqref{eq:logit_grad}, and RadAG constructs its reference by lowering the reward through exactly these logits.}
\rev{It first normalizes the sensitivity as}
\begin{equation}
\revcolor \tilde{\boldsymbol{g}}_{S_\ell}=\frac{\boldsymbol{g}_{S_\ell}-\operatorname{mean}_{\mathcal{K}}(\boldsymbol{g}_{S_\ell})}{\operatorname{RMS}(\boldsymbol{g}_{S_\ell})},
\label{eq:normalize}
\end{equation}
\rev{where $\operatorname{mean}_{\mathcal{K}}$ takes the mean of each row over the image keys $\mathcal{K}$ and $\operatorname{RMS}(\boldsymbol{x})=\|\boldsymbol{x}\|_2/\sqrt{\dim\boldsymbol{x}}$.}
\rev{The normalization serves two purposes.}
\rev{First, the centering removes the part of $\boldsymbol{g}_{S_\ell}$ that raises all image-key logits of a query equally; this part only shifts attention between image and text keys.}
\rev{Second, the division by the RMS makes the perturbation strength independent of the scale of the reward.}
\rev{The normalization is applied separately to image and text queries and leaves the logits to text keys unchanged (App.~\ref{app:proof}).}

\rev{RadAG then moves the logits against this direction and takes the resulting prediction as the reference,}
\begin{equation}
\revcolor \mathbf{S}_{\mathrm{adv}}=\mathbf{S}_\ell-\alpha\,\tilde{\boldsymbol{g}}_{S_\ell},\qquad
\boldsymbol{v}_{\mathrm{adv}}=\boldsymbol{v}_\theta(\boldsymbol{z}_t,t,c;\mathbf{S}_{\mathrm{adv}}),
\label{eq:perturbation}
\end{equation}
\rev{where $\alpha>0$ is the perturbation strength and $\boldsymbol{v}_\theta(\boldsymbol{z}_t,t,c;\mathbf{S})$ is the prediction with the logits of block $\ell$ replaced by $\mathbf{S}$, which is the base prediction $\boldsymbol{v}_\theta$ for $\mathbf{S}=\mathbf{S}_\ell$.}

\rev{Moving the logits against $\tilde{\boldsymbol{g}}_{S_\ell}$ lowers the reward most rapidly to first order, so $\boldsymbol{v}_{\mathrm{adv}}$ is a prediction with a lower target reward.}
\rev{Unlike gradient-based guidance, which differentiates the reward with respect to the latent through the entire generator, RadAG differentiates it only with respect to the attention logits on which it depends (Sec.~\ref{sec:motivation}), so the backward pass stops at block $\ell$ (Sec.~\ref{sec:efficient}).}
\rev{$\boldsymbol{v}_{\mathrm{adv}}$ thus plays the role of the unconditional prediction in CFG, and extrapolating away from it raises the reward.}

\paragraph{Update rule.}
\rev{RadAG replaces the reference prediction $\boldsymbol{v}_{\mathrm{ref}}$ in Eq.~\eqref{eq:extrapolation} with $\boldsymbol{v}_{\mathrm{adv}}$:}
\begin{equation}
\revcolor \boldsymbol{z}_{t'}=\boldsymbol{z}_t+(t'-t)\underbrace{\big[\boldsymbol{v}_\theta+w\big(\boldsymbol{v}_\theta-\boldsymbol{v}_{\mathrm{adv}}\big)\big]}_{\boldsymbol{v}_{\mathrm{RadAG}}},
\label{eq:radag_update}
\end{equation}
\rev{and repeats this step from $t=1$ to $t=0$ (Algorithm~\ref{alg:radag}).}

\paragraph{Reward ascent.}
\rev{To verify that moving away from $\boldsymbol{v}_{\mathrm{adv}}$ raises the reward, we write the reward of the clean estimate computed from a velocity $\boldsymbol{v}$ by Eq.~\eqref{eq:clean_estimate} as}
\begin{equation}
\revcolor R_t(\boldsymbol{v})=R\big(\mathcal{D}(\hat{\boldsymbol{z}}_{0|t}),c\big),\qquad\hat{\boldsymbol{z}}_{0|t}=\boldsymbol{z}_t-t\,\boldsymbol{v},
\label{eq:pred_reward}
\end{equation}
\rev{so that $\boldsymbol{g}_{S_\ell}=\nabla_{\mathbf{S}_\ell}R_t(\boldsymbol{v}_\theta)$.}
\rev{To first order in $\alpha$, the reference then lowers the reward by $\Delta=\alpha\langle\boldsymbol{g}_{S_\ell},\tilde{\boldsymbol{g}}_{S_\ell}\rangle\geq0$ (App.~\ref{app:proof_sign}), $R_t(\boldsymbol{v}_{\mathrm{adv}})=R_t(\boldsymbol{v}_\theta)-\Delta$, and Proposition~\ref{prop:ascent} shows that the update raises it.}

\begin{proposition}[Extrapolation from a reward-adversarial reference ascends the reward]
\label{prop:ascent}
\rev{Assume that $\mathbf{S}\mapsto\boldsymbol{v}_\theta(\boldsymbol{z}_t,t,c;\mathbf{S})$ is twice continuously differentiable near $\mathbf{S}_\ell$ and that $R_t$ is twice continuously differentiable near $\boldsymbol{v}_\theta$.}
For fixed $w>0$, as $\alpha\to0$,
\begin{equation}
\revcolor R_t(\boldsymbol{v}_{\mathrm{RadAG}})-R_t(\boldsymbol{v}_\theta)
=w\,\Delta+O(\alpha^2),
\label{eq:ascent}
\end{equation}
\textit{where $\Delta=\alpha\langle\boldsymbol{g}_{S_\ell},\tilde{\boldsymbol{g}}_{S_\ell}\rangle$.}
\end{proposition}
\rev{Proposition~\ref{prop:ascent} states that the update gains $w$ times the reward $\Delta$ that the reference loses, since by Eq.~\eqref{eq:radag_update} it reverses the deviation $\boldsymbol{v}_{\mathrm{adv}}-\boldsymbol{v}_\theta$ of the reference and scales it by $w$.}
\rev{Here $\langle\boldsymbol{g}_{S_\ell},\tilde{\boldsymbol{g}}_{S_\ell}\rangle$ is the rate at which the reward falls as the logits move along $-\tilde{\boldsymbol{g}}_{S_\ell}$, that is, the strength of the reward sensitivity of block $\ell$, so the gain grows with this sensitivity, which Sec.~\ref{sec:motivation} found to be large on the queried objects.}
\rev{Since $\Delta>0$ unless $\boldsymbol{g}_{S_\ell}$ is constant over the image keys of every query (App.~\ref{app:proof_sign}), RadAG raises the reward of the clean estimate for any differentiable target reward.}

\begin{figure}[t]
\centering
\includegraphics{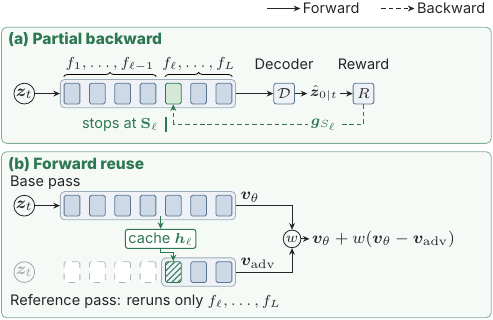}
\caption{\rev{\textbf{Efficient computation of RadAG.} (a) Partial backward: the reward gradient stops at $\mathbf{S}_\ell$. (b) Forward reuse: the reference pass reruns only $f_\ell,\dots,f_L$ from the cached $\boldsymbol{h}_\ell$ with $\mathbf{S}_{\mathrm{adv}}$.}}
\label{fig:efficiency}
\end{figure}

\begin{algorithm}[t]
\caption{Inference-Time Sampling with RadAG.}
\label{alg:radag}
\begin{algorithmic}[1]
\Require condition $c$, times $1=t_0>\dots>t_K=0$, block $\ell$, strengths $\alpha,w$
\State $\boldsymbol{z}_{t_0}\sim\mathcal{N}(\mathbf{0},\mathbf{I})$
\For{$k=0,\dots,K-1$}
\State $t\gets t_k$, $t'\gets t_{k+1}$
\State $\boldsymbol{h}_\ell,\boldsymbol{v}_\theta\gets$ Eq.~\eqref{eq:split} \Comment{graph from $\mathbf{S}_\ell$ on}
\State $\boldsymbol{g}_{S_\ell}\gets\nabla_{\mathbf{S}_\ell}R_t(\boldsymbol{v}_\theta)$ \Comment{partial backward}
\State $\mathbf{S}_{\mathrm{adv}}\gets\mathbf{S}_\ell-\alpha\tilde{\boldsymbol{g}}_{S_\ell}$ \Comment{Eq.~\eqref{eq:normalize}}
\State $\boldsymbol{v}_{\mathrm{adv}}\gets$ Eq.~\eqref{eq:split} from cached $\boldsymbol{h}_\ell$ \Comment{forward reuse}
\State $\boldsymbol{v}_{\mathrm{RadAG}}\gets\boldsymbol{v}_\theta+w(\boldsymbol{v}_\theta-\boldsymbol{v}_{\mathrm{adv}})$
\State $\boldsymbol{z}_{t'}\gets\boldsymbol{z}_t+(t'-t)\boldsymbol{v}_{\mathrm{RadAG}}$
\EndFor
\State \Return $\boldsymbol{z}_0$
\end{algorithmic}
\end{algorithm}
\subsection{Efficient Reward Gradient and Reference}
\label{sec:efficient}
\rev{Compared with existing extrapolation guidance, RadAG requires two additional computations at each step: a backward pass for $\boldsymbol{g}_{S_\ell}$, and a reference pass for $\boldsymbol{v}_{\mathrm{adv}}$, which needs $\boldsymbol{g}_{S_\ell}$ and therefore cannot run in the same batch as the base pass, unlike the reference passes of CFG and PAG~\cite{pag}.}
\rev{Omitting $t$ and $c$, we write the generator with $L$ blocks as $\boldsymbol{v}_\theta=f_L\circ\cdots\circ f_1(\boldsymbol{z}_t)$, let $\boldsymbol{h}_\ell=f_{\ell-1}\circ\cdots\circ f_1(\boldsymbol{z}_t)$ be the input of block $\ell$, and let $f_\ell(\boldsymbol{h}_\ell;\mathbf{S})$ denote block $\ell$ with its attention logits set to $\mathbf{S}$.}
Then
\begin{equation}
\begin{aligned}\boldsymbol{v}_\theta&=f_L\circ\cdots\circ f_\ell(\boldsymbol{h}_\ell;\mathbf{S}_\ell),\\
\boldsymbol{v}_{\mathrm{adv}}&=f_L\circ\cdots\circ f_\ell(\boldsymbol{h}_\ell;\mathbf{S}_{\mathrm{adv}}),\end{aligned}
\label{eq:split}
\end{equation}
\rev{where $\boldsymbol{h}_\ell$ depends on neither $\mathbf{S}_\ell$ nor $\mathbf{S}_{\mathrm{adv}}$. Fig.~\ref{fig:efficiency} shows how RadAG uses this split in two ways.}

\paragraph{Partial backward.}
\rev{Since $\boldsymbol{h}_\ell$ does not depend on $\mathbf{S}_\ell$, the reward sensitivity $\boldsymbol{g}_{S_\ell}=\nabla_{\mathbf{S}_\ell}R_t(\boldsymbol{v}_\theta)$ depends only on the computation after $\mathbf{S}_\ell$.}
\rev{RadAG therefore computes $\boldsymbol{h}_\ell$ without storing its computation graph, so the backward pass stops at block $\ell$ instead of traversing the entire generator as for $\boldsymbol{g}_{z_t}$ (Fig.~\ref{fig:efficiency}a).}
\rev{Block $\ell$ keeps the fused attention kernel~\cite{flashattention} in the forward pass and recomputes $\mathbf{S}_\ell$ only for the backward pass (App.~\ref{app:ste}).}

\paragraph{Forward reuse.}
\rev{The base and reference passes differ only from block $\ell$ on, so they share $\boldsymbol{h}_\ell$.}
\rev{RadAG therefore caches $\boldsymbol{h}_\ell$ in the base pass and reruns only $f_\ell,\dots,f_L$ with $\mathbf{S}_{\mathrm{adv}}$ (Fig.~\ref{fig:efficiency}b).}
\rev{Each step thus needs one full forward pass, one forward pass of $f_\ell,\dots,f_L$, and one backward pass from the reward to block $\ell$, and a later block shortens both partial passes (Sec.~\ref{sec:ablation}); Algorithm~\ref{alg:radag} summarizes the procedure.}

\begin{table*}[t]
\centering
\captionof{table}{\textbf{Reward alignment across model scales.} Mean scores over 5,000 COCO prompts; \enabled marks the target reward. Bold and underline mark the best and second-best guided results, ranked on unrounded scores.}
\label{tab:main}
\begingroup
\fontsize{8}{9.5}\selectfont
\setlength{\tabcolsep}{2.0pt}
\renewcommand{\arraystretch}{0.98}
\begin{tabularx}{\textwidth}{@{}c@{\hspace{1pt}}c@{\hspace{3pt}}l*{8}{>{\centering\arraybackslash}X}}
\toprule
& & & \multicolumn{4}{c}{\textbf{FLUX.2-klein-4B}} & \multicolumn{4}{c}{\textbf{FLUX.2-klein-9B}}\\
\cmidrule(lr){4-7}\cmidrule(lr){8-11}
\multicolumn{11}{@{}l}{\textbf{(a) Text alignment: PickScore guidance}}\\[2pt]
& & Method & \textcolor{targetgreen}{Pick$\uparrow$\,$^{\text{\enabled}}$} & CLIP$\uparrow$ & IR$\uparrow$ & HPSv2$\uparrow$ & \textcolor{targetgreen}{Pick$\uparrow$\,$^{\text{\enabled}}$} & CLIP$\uparrow$ & IR$\uparrow$ & HPSv2$\uparrow$\\
\cmidrule(r){3-3}\cmidrule(lr){4-7}\cmidrule(l){8-11}
\multirow{4}{*}{\rotatebox{90}{\scriptsize\color{tablegray} Reward-free}} & \ldelim\{{4}{3.2mm} & Vanilla & 22.893 & 26.462 & 1.071 & 0.296 & 23.072 & 26.737 & 1.207 & 0.301\\
 &  & SAG~\cite{sag} & 22.836 & 26.422 & 1.079 & 0.297 & 22.976 & 26.699 & 1.206 & 0.302\\
 &  & PAG~\cite{pag} & 22.374 & \textbf{26.961} & 1.093 & 0.297 & 22.871 & \textbf{26.973} & 1.206 & 0.296\\
 &  & SEG~\cite{seg} & 21.741 & \underline{26.941} & 0.993 & 0.282 & 22.794 & 26.578 & 1.209 & 0.299\\
\addlinespace[4pt]
\multirow{6}{*}{\rotatebox{90}{\scriptsize\color{tablegray} Reward-based}} & \ldelim\{{6}{3.2mm} & DPS~\cite{dps} & 23.034 & 26.506 & 1.087 & 0.297 & 23.208 & 26.750 & 1.223 & 0.303\\
 &  & FreeDoM~\cite{freedom} & 24.205 & 26.659 & 1.107 & 0.309 & 23.997 & 26.809 & 1.207 & 0.310\\
 &  & SVDD-PM~\cite{svdd} & 23.390 & 26.713 & 1.138 & 0.302 & 23.598 & \underline{26.921} & 1.261 & 0.309\\
 &  & NoiseTilt~\cite{noisetilt} & 24.946 & 26.601 & 1.159 & 0.316 & 24.649 & 26.787 & 1.261 & 0.318\\
\cmidrule(r){3-3}\cmidrule(lr){4-7}\cmidrule(l){8-11}
 &  & \cellcolor{ourshighlight}\textbf{RadAG} (Ours) & \cellcolor{ourshighlight}\underline{25.293} & \cellcolor{ourshighlight}26.765 & \cellcolor{ourshighlight}\underline{1.195} & \cellcolor{ourshighlight}\underline{0.321} & \cellcolor{ourshighlight}\underline{25.477} & \cellcolor{ourshighlight}26.857 & \cellcolor{ourshighlight}\underline{1.300} & \cellcolor{ourshighlight}\underline{0.325}\\
 &  & \cellcolor{ourshighlight}\textbf{RadAG+} (Ours) & \cellcolor{ourshighlight}\textbf{25.831} & \cellcolor{ourshighlight}26.693 & \cellcolor{ourshighlight}\textbf{1.216} & \cellcolor{ourshighlight}\textbf{0.326} & \cellcolor{ourshighlight}\textbf{25.932} & \cellcolor{ourshighlight}26.871 & \cellcolor{ourshighlight}\textbf{1.314} & \cellcolor{ourshighlight}\textbf{0.330}\\
\addlinespace[3pt]
\midrule
\addlinespace[2pt]
\multicolumn{11}{@{}l}{
\textbf{(b) Aesthetic quality: Aesthetic guidance}}\\[2pt]
& & Method & \textcolor{targetgreen}{AES v2$\uparrow$\,$^{\text{\enabled}}$} & Q-Align$\uparrow$ & NIMA$\uparrow$ & IR$\uparrow$ & \textcolor{targetgreen}{AES v2$\uparrow$\,$^{\text{\enabled}}$} & Q-Align$\uparrow$ & NIMA$\uparrow$ & IR$\uparrow$\\
\cmidrule(r){3-3}\cmidrule(lr){4-7}\cmidrule(l){8-11}
\multirow{4}{*}{\rotatebox{90}{\scriptsize\color{tablegray} Reward-free}} & \ldelim\{{4}{3.2mm} & Vanilla & 5.292 & 2.871 & 5.125 & 1.071 & 5.396 & 3.070 & 5.205 & 1.207\\
 &  & SAG~\cite{sag} & 5.299 & 2.920 & 5.215 & 1.078 & 5.402 & 3.084 & 5.221 & 1.210\\
 &  & PAG~\cite{pag} & 5.426 & 3.009 & 5.265 & \underline{1.114} & 5.427 & 3.266 & 5.279 & 1.217\\
 &  & SEG~\cite{seg} & 5.464 & 3.101 & 5.323 & \textbf{1.145} & 5.450 & \textbf{3.470} & 5.340 & 1.209\\
\addlinespace[4pt]
\multirow{6}{*}{\rotatebox{90}{\scriptsize\color{tablegray} Reward-based}} & \ldelim\{{6}{3.2mm} & DPS~\cite{dps} & 5.335 & 2.881 & 5.132 & 1.073 & 5.432 & 3.082 & 5.212 & 1.208\\
 &  & FreeDoM~\cite{freedom} & 6.149 & \underline{3.132} & \underline{5.408} & 1.026 & 6.085 & 3.238 & 5.392 & 1.149\\
 &  & SVDD-PM~\cite{svdd} & 5.570 & 3.040 & 5.241 & 1.079 & 5.666 & 3.226 & 5.300 & 1.210\\
 &  & NoiseTilt~\cite{noisetilt} & 6.376 & 3.061 & 5.312 & 1.095 & 6.255 & 3.203 & 5.327 & 1.215\\
\cmidrule(r){3-3}\cmidrule(lr){4-7}\cmidrule(l){8-11}
 &  & \cellcolor{ourshighlight}\textbf{RadAG} (Ours) & \cellcolor{ourshighlight}\underline{6.521} & \cellcolor{ourshighlight}3.110 & \cellcolor{ourshighlight}5.328 & \cellcolor{ourshighlight}1.107 & \cellcolor{ourshighlight}\underline{6.621} & \cellcolor{ourshighlight}3.362 & \cellcolor{ourshighlight}\underline{5.422} & \cellcolor{ourshighlight}\underline{1.223}\\
 &  & \cellcolor{ourshighlight}\textbf{RadAG+} (Ours) & \cellcolor{ourshighlight}\textbf{6.726} & \cellcolor{ourshighlight}\textbf{3.211} & \cellcolor{ourshighlight}\textbf{5.420} & \cellcolor{ourshighlight}1.099 & \cellcolor{ourshighlight}\textbf{6.799} & \cellcolor{ourshighlight}\underline{3.397} & \cellcolor{ourshighlight}\textbf{5.484} & \cellcolor{ourshighlight}\textbf{1.227}\\
\bottomrule
\end{tabularx}
\endgroup
\end{table*}

\subsection{Extrapolation Refinements: RadAG+}
\label{sec:combine}
\rev{Our update rule in Eq.~\eqref{eq:radag_update} keeps the form of Eq.~\eqref{eq:extrapolation} with $\boldsymbol{v}_{\mathrm{adv}}$ as the negative branch, so techniques developed to refine extrapolation guidance also apply to RadAG, and RadAG+ adopts two of them.}

\rev{First, attention-level guidance such as NAG~\cite{nag} and PLADIS~\cite{pladis} extrapolates between the attention outputs of a positive and a negative branch rather than between final predictions.}
\rev{Gradient-based and candidate-based reward guidance has no negative attention branch, but the reference pass of RadAG already computes one, the attention output of the reward-adversarial logits, $\mathrm{Attn}_{\mathrm{adv}}=\texttt{Softmax}(\mathbf{S}_{\mathrm{adv}})\mathbf{V}_\ell$.}
\rev{RadAG+ therefore extrapolates}
\begin{equation}
\begin{aligned}\widetilde{\mathrm{Attn}}&=\mathrm{Attn}_\ell+\gamma\big(\mathrm{Attn}_\ell-\mathrm{Attn}_{\mathrm{adv}}\big),\\
\boldsymbol{v}_{\mathrm{attn}}&=f_L\circ\cdots\circ f_\ell(\boldsymbol{h}_\ell;\mathrm{Attn}_+),\end{aligned}
\label{eq:vattn}
\end{equation}
\rev{where $\gamma>0$, $\mathrm{Attn}_+$ is $\widetilde{\mathrm{Attn}}$ after the norm bound and blend of NAG (App.~\ref{app:radagplus}), and block $\ell$ uses $\mathrm{Attn}_+$ as its attention output.}
\rev{$\mathrm{Attn}_{\mathrm{adv}}$ comes from the reference pass at no cost, and computing $\boldsymbol{v}_{\mathrm{attn}}$ takes one more forward pass of $f_\ell,\dots,f_L$ from the cached $\boldsymbol{h}_\ell$.}

\rev{Second, as CFG++~\cite{cfgpp} does for CFG, RadAG+ uses the guided velocity only to predict the clean sample and keeps $\boldsymbol{v}_\theta$ to predict the noise that the update adds back to the latent.}
\rev{The guidance thus changes the predicted clean sample without altering the predicted noise, which keeps the latent close to those the generator was trained on; this needs a guided prediction in the form of Eq.~\eqref{eq:extrapolation}, which gradient-based and candidate-based reward guidance do not form.}
\rev{With both refinements, Eq.~\eqref{eq:radag_update} becomes}
\begin{equation}
\revcolor \boldsymbol{z}_{t'}=\boldsymbol{z}_t+(t'-t)\underbrace{\Big[\boldsymbol{v}_\theta+\kappa_t\Big(w\big(\boldsymbol{v}_\theta-\boldsymbol{v}_{\mathrm{adv}}\big)+\boldsymbol{v}_{\mathrm{attn}}-\boldsymbol{v}_\theta\Big)\Big]}_{\boldsymbol{v}_{\mathrm{RadAG+}}},
\label{eq:nag_guided}
\end{equation}
\rev{where $\kappa_t=t(1-t')/(t-t')$ results from the noise estimate (App.~\ref{app:radagplus}).}
\rev{With $\kappa_t=1$ and without $\boldsymbol{v}_{\mathrm{attn}}-\boldsymbol{v}_\theta$, Eq.~\eqref{eq:nag_guided} reduces to Eq.~\eqref{eq:radag_update}.}

\section{Experiments}
\label{sec:experiments}

\subsection{Setup}
\label{sec:setup}
\paragraph{Models and data.}
We evaluate FLUX.2-klein-4B and FLUX.2-klein-9B~\cite{flux2} at $512^2$ with four sampling steps and without CFG on COCO val2014 captions~\cite{coco}.
\rev{For $\boldsymbol{\epsilon}$-prediction models such as SDXL, RadAG uses the corresponding clean estimate in place of Eq.~\eqref{eq:clean_estimate} (Sec.~\ref{app:sd3}).}
Table~\ref{tab:main} and Fig.~\ref{fig:reward_time} use 5,000 prompts; all other quantitative evaluations use 1,000 prompts.
All methods share the evaluation captions and initial noise seeds, and hyperparameters are tuned on 100 separate captions (Sec.~\ref{app:protocol}).

\paragraph{Target and held-out rewards.}
We guide each model with PickScore~\cite{pickscore} or the LAION aesthetic predictor~\cite{laionaes}.
We call the reward used for guidance the target reward and the evaluation rewards not used for guidance the held-out rewards.
The held-out rewards are CLIP score~\cite{clipscore}, ImageReward~\cite{imagereward}, and HPSv2~\cite{hpsv2} for PickScore guidance, and Q-Align~\cite{qalign}, NIMA~\cite{nima}, and ImageReward for guidance with the aesthetic predictor.

\paragraph{Baselines.}
\rev{We compare RadAG and RadAG+ against the unguided model (Vanilla), perturbation-based attention guidance (SAG~\cite{sag}, PAG~\cite{pag}, SEG~\cite{seg}), gradient-based guidance (DPS~\cite{dps}, FreeDoM~\cite{freedom}, NoiseTilt~\cite{noisetilt}), and candidate-based search (SVDD-PM~\cite{svdd}).}
\rev{SAG, PAG, and SEG use the block of RadAG, and all baselines keep their original settings (Sec.~\ref{app:baselines}).}

\subsection{Main Results}
\label{sec:main}
\paragraph{Quantitative results.}
\rev{To test whether RadAG raises the target reward at a lower cost than existing inference-time guidance, Table~\ref{tab:main} reports the target and held-out rewards.}
\rev{RadAG attains the highest target reward among the baselines in all four settings.}
\rev{This gain does not come at the expense of other rewards: RadAG ranks among the top three on all twelve held-out rewards, and it scores higher than the baseline in 75 of the 84 combinations of the seven baselines and the twelve held-out rewards.}
\rev{Perturbation-based attention guidance barely moves the target reward with its heuristic references, and PAG and SEG even lower PickScore below Vanilla.}
\rev{RadAG+ raises the target reward further in all four settings and improves ten of the twelve held-out rewards over RadAG.}

\paragraph{Qualitative results.}
\rev{Fig.~\ref{fig:teaser} compares samples of Vanilla, DPS, NoiseTilt, and RadAG, and App.~\ref{app:qual} shows more samples on both scales and on SD3 Medium.}
\rev{On SD3 Medium with CFG, RadAG attains the second-highest PickScore after FreeDoM and the highest CLIP score in 2.3 times less time than FreeDoM (Sec.~\ref{app:sd3}).}

\paragraph{Performance--time trade-off.}
\rev{Fig.~\ref{fig:reward_time} places these rewards against the time per image.}
\rev{RadAG takes 30 to 65\% less time than the gradient-based and candidate-based baselines and lies at the upper left in every setting; Sec.~\ref{sec:cost} analyzes where this saving comes from.}

\subsection{Inference Cost}
\label{sec:cost}
\begin{figure}[!htbp]
\centering
\includegraphics[width=\linewidth]{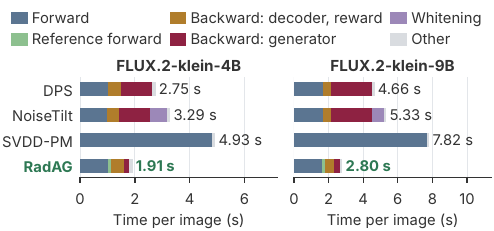}
\caption{\rev{\textbf{Time per image of reward-based methods by stage.} PickScore guidance, 1,000 prompts, one NVIDIA RTX A6000, summed over four steps. Forward includes the generator, decoder, and reward model.}}
\label{fig:cost}
\end{figure}

\rev{To see where the saving in Fig.~\ref{fig:reward_time} comes from, Fig.~\ref{fig:cost} breaks the time per image into stages.}
\rev{RadAG and DPS both take one reward gradient per step and share every other stage, so they differ mainly in the backward pass through the generator.}
\rev{Since this pass stops at block $\ell$, RadAG makes it 6.1 and 6.5 times faster on the 4B and 9B models, while its reference forward pass adds only 5 to 7\% of the total time, so RadAG takes 30 to 40\% less time than DPS.}
\rev{NoiseTilt and SVDD-PM are slower than DPS owing to gradient whitening and candidate passes (Sec.~\ref{app:breakdown}).}

\subsection{Ablation Study}
\label{sec:ablation}
\rev{We ablate where RadAG modifies the generator, from coarse to fine: the block (Table~\ref{tab:depth}), the variable within the block (Table~\ref{tab:location}), and the region of the logits (Table~\ref{tab:region}). Tables~\ref{tab:depth}--\ref{tab:region} use PickScore guidance on FLUX.2-klein-4B.}

\paragraph{Block depth and forward reuse.}
\rev{Table~\ref{tab:depth} moves block $\ell$ from early to late, with and without forward reuse.}
\rev{The later the block, the fewer blocks the backward pass traverses, so the time decreases in both groups.}
\rev{With forward reuse, the reference pass also reruns fewer blocks, so the time decreases faster and the gap between the two groups widens.}
\rev{Later blocks also give higher PickScore and ImageReward, so the latest block is the cheapest and the most effective.}

\begin{table}[t]
\centering
\renewcommand{\arraystretch}{1.32}
\caption{\rev{\textbf{Block depth and forward reuse.} Blocks counts block $\ell$ and the later ones out of 25; time is seconds per image.}}
\label{tab:depth}
\footnotesize\setlength{\tabcolsep}{2pt}
\begin{tabularx}{\linewidth}{@{}c@{\hspace{1pt}}c@{\hspace{3pt}}l*{5}{>{\raggedleft\arraybackslash}X}@{}}
\toprule
& & Blocks & Pick$\uparrow$ & CLIP$\uparrow$ & IR$\uparrow$ & HPSv2$\uparrow$ & Time$\downarrow$\\
\cmidrule(r){3-3}\cmidrule(l){4-8}
& & Vanilla & 22.891 & 26.416 & 1.059 & 0.296 & \textbf{0.611}\\
\cmidrule(r){3-3}\cmidrule(l){4-8}
\multirow{3}{*}{\rotatebox{90}{\scriptsize\color{tablegray} Without reuse}} & \ldelim\{{3}{3.2mm} & 13 (early) & 24.508 & 26.692 & 1.146 & 0.315 & 2.690\\
& & 7 (middle) & 24.996 & \textbf{26.857} & 1.179 & 0.320 & 2.410\\
& & 4 (late) & 25.280 & 26.759 & 1.182 & \textbf{0.321} & 2.256\\
\cmidrule(r){3-3}\cmidrule(l){4-8}
\multirow{3}{*}{\rotatebox{90}{\scriptsize\color{tablegray} With reuse}} & \ldelim\{{3}{3.2mm} & 13 (early) & 24.508 & 26.723 & 1.136 & 0.315 & 2.445\\
& & 7 (middle) & 24.998 & 26.849 & 1.183 & 0.320 & 2.045\\
& & \cellcolor{ourshighlight}4 (late) & \cellcolor{ourshighlight}\textbf{25.290} & \cellcolor{ourshighlight}26.797 & \cellcolor{ourshighlight}\textbf{1.186} & \cellcolor{ourshighlight}\textbf{0.321} & \cellcolor{ourshighlight}1.864\\
\bottomrule
\end{tabularx}
\end{table}

\begin{table}[!htbp]
\centering
\renewcommand{\arraystretch}{1.05}
\caption{\rev{\textbf{Modification location.} Each location uses its own tuned strength and guidance scale; time is seconds per image.}}
\label{tab:location}
\footnotesize\setlength{\tabcolsep}{2pt}
\begin{tabularx}{\linewidth}{l*{5}{>{\raggedleft\arraybackslash}X}}
\toprule
Location & Pick$\uparrow$ & CLIP$\uparrow$ & IR$\uparrow$ & HPSv2$\uparrow$ & Time$\downarrow$\\
\cmidrule(r){1-1}\cmidrule(l){2-6}
Vanilla & 22.891 & 26.420 & 1.059 & 0.296 & \textbf{0.611}\\
\cmidrule(r){1-1}\cmidrule(l){2-6}
Value $\mathbf{V}_\ell$ & 25.066 & \textbf{26.902} & \textbf{1.194} & 0.315 & 1.768\\
Map $\mathbf{A}_\ell$ & 25.356 & 26.840 & 1.175 & \textbf{0.323} & 1.848\\
Output $\mathrm{Attn}_\ell$ & \textbf{25.425} & 26.862 & 1.161 & 0.321 & 1.761\\
Block output & 24.490 & 26.632 & 1.118 & 0.308 & 1.736\\
\oursrow
Logits $\mathbf{S}_\ell$ & 25.290 & 26.770 & 1.186 & 0.321 & 1.864\\
\bottomrule
\end{tabularx}
\end{table}

\paragraph{Modification location.}
\rev{To test whether the gain is specific to the logits, Table~\ref{tab:location} applies the same construction to other variables of block $\ell$ (Sec.~\ref{app:locations}).}
\rev{Every variable inside the attention raises the target reward well above Vanilla, while the block output raises it less, so the gain comes from the reward-adversarial reference within the attention rather than a particular variable.}
\rev{We use the logits, on which Sec.~\ref{sec:motivation} and Eq.~\eqref{eq:normalize} are defined.}

\begin{table}[t]
\centering
\renewcommand{\arraystretch}{1.05}
\caption{\rev{\textbf{Attention region.} Letters give the query and key modalities, image (I) or text (T).}}
\label{tab:region}
\footnotesize\setlength{\tabcolsep}{2pt}
\begin{tabularx}{\linewidth}{X*{4}{>{\raggedleft\arraybackslash}X}}
\toprule
Region & Pick$\uparrow$ & CLIP$\uparrow$ & IR$\uparrow$ & HPSv2$\uparrow$\\
\cmidrule(r){1-1}\cmidrule(l){2-5}
Vanilla & 22.891 & 26.420 & 1.059 & 0.296\\
\cmidrule(r){1-1}\cmidrule(l){2-5}
II & 25.167 & 26.760 & 1.180 & \textbf{0.321}\\
TI & 23.361 & 26.490 & 1.076 & 0.300\\
IT & 22.673 & \textbf{27.020} & 1.055 & 0.293\\
TT & 23.143 & 26.550 & 1.063 & 0.297\\
\oursrow
II+TI & \textbf{25.290} & 26.770 & \textbf{1.186} & \textbf{0.321}\\
\bottomrule
\end{tabularx}
\end{table}

\paragraph{Attention region.}
\rev{To find which logits carry the gain, Table~\ref{tab:region} modifies each query--key region of the joint attention separately, where II denotes the logits from image queries to image keys, TI from text to image, IT from image to text, and TT from text to text.} 
\rev{The logits to image keys (II and TI) carry most of the gain, and modifying both gives the highest PickScore and ImageReward, while the logits to text keys barely help and IT even falls below Vanilla, so RadAG modifies only the logits to image keys.}

\begin{table}[t]
\centering
\caption{\rev{\textbf{Components of RadAG+.} Attention-output extrapolation (Attn.) and separate noise estimate (Noise). $\Delta$Time is relative to RadAG; all four settings are in Table~\ref{tab:plus_full}.}}
\label{tab:plus}
\footnotesize\setlength{\tabcolsep}{2.5pt}
\begin{tabularx}{\linewidth}{@{}cc*{5}{>{\centering\arraybackslash}X}@{}}
\toprule
\multicolumn{7}{@{}l}{\textbf{(a) FLUX.2-klein-4B, PickScore guidance}}\\
Attn. & Noise & Pick$\uparrow$ & CLIP$\uparrow$ & IR$\uparrow$ & HPSv2$\uparrow$ & $\Delta$Time\\
\cmidrule(r){1-2}\cmidrule(l){3-7}
 &  & 25.282 & 26.772 & 1.184 & 0.321 & --\\
 & \ding{51} & 25.420 & \textbf{26.773} & 1.186 & 0.321 & $-$1\%\\
\ding{51} &  & 25.547 & 26.690 & 1.175 & 0.322 & +6\%\\
\oursrow \ding{51} & \ding{51} & \textbf{25.746} & 26.641 & \textbf{1.188} & \textbf{0.324} & +5\%\\
\midrule
\multicolumn{7}{@{}l}{\textbf{(b) FLUX.2-klein-9B, aesthetic guidance}}\\
Attn. & Noise & AES$\uparrow$ & Q-Align$\uparrow$ & NIMA$\uparrow$ & IR$\uparrow$ & $\Delta$Time\\
\cmidrule(r){1-2}\cmidrule(l){3-7}
 &  & 6.621 & 3.346 & 5.411 & 1.236 & --\\
 & \ding{51} & 6.712 & \textbf{3.390} & 5.453 & 1.240 & 0\%\\
\ding{51} &  & 6.701 & 3.347 & 5.436 & \textbf{1.243} & +8\%\\
\oursrow \ding{51} & \ding{51} & \textbf{6.782} & 3.378 & \textbf{5.479} & 1.240 & +8\%\\
\bottomrule
\end{tabularx}
\end{table}

\paragraph{Components of RadAG+.}
\rev{RadAG alone already attains the highest target reward among the baselines (Table~\ref{tab:main}), and RadAG+ raises it further at almost no cost.}
\rev{Table~\ref{tab:plus} adds each refinement of Sec.~\ref{sec:combine} to RadAG separately: each raises the target reward in all four settings (Table~\ref{tab:plus_full}), and together they give the highest, close to the sum of the separate gains, while only the attention-output extrapolation adds time (5 to 8\%).}

\begin{figure}[t]
\centering
\includegraphics[width=\linewidth]{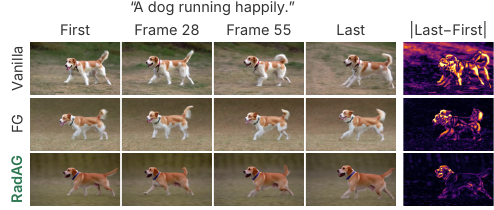}
\caption{\rev{\textbf{Looped video generation on Wan2.1-T2V-1.3B.} $|\mathrm{Last}-\mathrm{First}|$ is the per-pixel difference of the last and first frames; darker means a more seamless loop. FG: Frame Guidance~\cite{frameguidance}.}}
\label{fig:video}
\end{figure}

\subsection{Extension to Video Generation}
\label{sec:video}
\rev{To test whether RadAG extends to video generation, we align Wan2.1-T2V-1.3B~\cite{wan} to a loop reward that compares the first and last frames and compare RadAG with Frame Guidance~\cite{frameguidance}, which backpropagates this reward through the entire generator up to seven times per step, whereas RadAG takes one gradient per step at the 21st of 30 blocks (App.~\ref{app:video}).}
\rev{RadAG closes the loop 3.4 times faster, and Fig.~\ref{fig:video} shows that it keeps the scene in motion while matching the first and last frames.}

\section{Conclusion}
\label{sec:conclusion}
\rev{Extrapolation guidance is cheap but cannot steer sampling toward a reward, whereas reward guidance steers toward a reward but backpropagates through the entire generator.}
\rev{RadAG bridges the two: it builds the reference of extrapolation guidance from the reward sensitivity of attention, so that moving away from it provably raises the reward, and computes it with a backward pass that stops at one block.}
\rev{Across two FLUX.2-klein scales and two rewards, RadAG attains the highest target reward with 30--65\% less time than gradient-based and candidate-based baselines, and it extends to video generation.}
\rev{Needing only the attention logits of one block, RadAG turns any differentiable reward into guidance for attention-based image and video generators at nearly the cost of extrapolation guidance.}

{
    \small
    \bibliographystyle{ieeenat_fullname}
    \bibliography{main}

\begin{thebibliography}{69}
\providecommand{\natexlab}[1]{#1}
\providecommand{\url}[1]{\texttt{#1}}
\expandafter\ifx\csname urlstyle\endcsname\relax
  \providecommand{\doi}[1]{doi: #1}\else
  \providecommand{\doi}{doi: \begingroup \urlstyle{rm}\Url}\fi

\bibitem[Ahn et~al.(2024)Ahn, Cho, Min, Jang, Kim, Kim, Park, Jin, and
  Kim]{pag}
Donghoon Ahn, Hyoungwon Cho, Jaewon Min, Wooseok Jang, Jungwoo Kim, SeonHwa
  Kim, Hyun~Hee Park, Kyong~Hwan Jin, and Seungryong Kim.
\newblock {Self-Rectifying Diffusion Sampling with Perturbed-Attention
  Guidance}.
\newblock In \emph{Proceedings of the European Conference on Computer Vision
  (ECCV)}, pages 1--17, 2024.

\bibitem[Ahn et~al.(2025)Ahn, Kang, Lee, Kim, Jang, Min, Lee, Paul, and
  Kim]{headhunter}
Donghoon Ahn, Jiwon Kang, Sanghyun Lee, Minjae Kim, Wooseok Jang, Jaewon Min,
  Sangwu Lee, Sayak Paul, and Seungryong Kim.
\newblock {Where and How to Perturb: On the Design of Perturbation Guidance in
  Diffusion and Flow Models}.
\newblock In \emph{Advances in Neural Information Processing Systems
  (NeurIPS)}, pages 128348--128396, 2025.

\bibitem[Arras et~al.(2022)Arras, Osman, and Samek]{clevrxai}
Leila Arras, Ahmed Osman, and Wojciech Samek.
\newblock {CLEVR-XAI: A Benchmark Dataset for the Ground Truth Evaluation of
  Neural Network Explanations}.
\newblock \emph{Information Fusion}, 81:\penalty0 14--40, 2022.

\bibitem[Bansal et~al.(2024)Bansal, Chu, Schwarzschild, Sengupta, Goldblum,
  Geiping, and Goldstein]{universal}
Arpit Bansal, Hong-Min Chu, Avi Schwarzschild, Soumyadip Sengupta, Micah
  Goldblum, Jonas Geiping, and Tom Goldstein.
\newblock {Universal Guidance for Diffusion Models}.
\newblock In \emph{International Conference on Learning Representations
  (ICLR)}, pages 51304--51323, 2024.

\bibitem[Ben-Hamu et~al.(2024)Ben-Hamu, Puny, Gat, Karrer, Singer, and
  Lipman]{dflow}
Heli Ben-Hamu, Omri Puny, Itai Gat, Brian Karrer, Uriel Singer, and Yaron
  Lipman.
\newblock {D-Flow: Differentiating through Flows for Controlled Generation}.
\newblock In \emph{International Conference on Machine Learning (ICML)}, pages
  3462--3483, 2024.

\bibitem[Bengio et~al.(2013)Bengio, L{\'e}onard, and Courville]{ste}
Yoshua Bengio, Nicholas L{\'e}onard, and Aaron Courville.
\newblock {Estimating or Propagating Gradients Through Stochastic Neurons for
  Conditional Computation}.
\newblock arXiv preprint arXiv:1308.3432, 2013.

\bibitem[Black et~al.(2024)Black, Janner, Du, Kostrikov, and Levine]{ddpo}
Kevin Black, Michael Janner, Yilun Du, Ilya Kostrikov, and Sergey Levine.
\newblock {Training Diffusion Models with Reinforcement Learning}.
\newblock In \emph{International Conference on Learning Representations
  (ICLR)}, pages 4965--4987, 2024.

\bibitem[{Black Forest Labs}(2026)]{flux2}
{Black Forest Labs}.
\newblock {FLUX.2 [klein]: Towards Interactive Visual Intelligence}.
\newblock
  \url{https://bfl.ai/blog/flux2-klein-towards-interactive-visual-intelligence},
  2026.
\newblock Black Forest Labs blog post, January 15, 2026.

\bibitem[Chefer et~al.(2023)Chefer, Alaluf, Vinker, Wolf, and
  Cohen-Or]{attendexcite}
Hila Chefer, Yuval Alaluf, Yael Vinker, Lior Wolf, and Daniel Cohen-Or.
\newblock {Attend-and-Excite: Attention-Based Semantic Guidance for
  Text-to-Image Diffusion Models}.
\newblock \emph{ACM Transactions on Graphics (SIGGRAPH)}, 42\penalty0
  (4):\penalty0 148:1--148:10, 2023.

\bibitem[Chen et~al.(2025)Chen, Bandyopadhyay, Zou, and Song]{nag}
Dar-Yen Chen, Hmrishav Bandyopadhyay, Kai Zou, and Yi-Zhe Song.
\newblock {Normalized Attention Guidance: Universal Negative Guidance for
  Diffusion Models}.
\newblock In \emph{Advances in Neural Information Processing Systems
  (NeurIPS)}, pages 137412--137446, 2025.

\bibitem[Cheng et~al.(2022)Cheng, Misra, Schwing, Kirillov, and
  Girdhar]{mask2former}
Bowen Cheng, Ishan Misra, Alexander~G. Schwing, Alexander Kirillov, and Rohit
  Girdhar.
\newblock {Masked-attention Mask Transformer for Universal Image Segmentation}.
\newblock In \emph{Proceedings of the IEEE/CVF Conference on Computer Vision
  and Pattern Recognition (CVPR)}, pages 1290--1299, 2022.

\bibitem[Chung et~al.(2023)Chung, Kim, McCann, Klasky, and Ye]{dps}
Hyungjin Chung, Jeongsol Kim, Michael~Thompson McCann, Marc~Louis Klasky, and
  Jong~Chul Ye.
\newblock {Diffusion Posterior Sampling for General Noisy Inverse Problems}.
\newblock In \emph{International Conference on Learning Representations
  (ICLR)}, 2023.

\bibitem[Chung et~al.(2025)Chung, Kim, Park, Nam, and Ye]{cfgpp}
Hyungjin Chung, Jeongsol Kim, Geon~Yeong Park, Hyelin Nam, and Jong~Chul Ye.
\newblock {CFG++: Manifold-constrained Classifier Free Guidance for Diffusion
  Models}.
\newblock In \emph{International Conference on Learning Representations
  (ICLR)}, pages 30824--30850, 2025.

\bibitem[Clark et~al.(2024)Clark, Vicol, Swersky, and Fleet]{draft}
Kevin Clark, Paul Vicol, Kevin Swersky, and David~J. Fleet.
\newblock {Directly Fine-Tuning Diffusion Models on Differentiable Rewards}.
\newblock In \emph{International Conference on Learning Representations
  (ICLR)}, pages 4793--4822, 2024.

\bibitem[Dao et~al.(2022)Dao, Fu, Ermon, Rudra, and R{\'e}]{flashattention}
Tri Dao, Daniel~Y. Fu, Stefano Ermon, Atri Rudra, and Christopher R{\'e}.
\newblock {FlashAttention: Fast and Memory-Efficient Exact Attention with
  IO-Awareness}.
\newblock In \emph{Advances in Neural Information Processing Systems
  (NeurIPS)}, pages 16344--16359, 2022.

\bibitem[Dhariwal and Nichol(2021)]{cg}
Prafulla Dhariwal and Alexander Nichol.
\newblock {Diffusion Models Beat GANs on Image Synthesis}.
\newblock In \emph{Advances in Neural Information Processing Systems}, pages
  8780--8794, 2021.

\bibitem[Domingo-Enrich et~al.(2025)Domingo-Enrich, Drozdzal, Karrer, and
  Chen]{adjointmatching}
Carles Domingo-Enrich, Michal Drozdzal, Brian Karrer, and Ricky T.~Q. Chen.
\newblock {Adjoint Matching: Fine-tuning Flow and Diffusion Generative Models
  with Memoryless Stochastic Optimal Control}.
\newblock In \emph{International Conference on Learning Representations
  (ICLR)}, pages 53791--53846, 2025.

\bibitem[Efron(2011)]{efron2011tweedie}
Bradley Efron.
\newblock {Tweedie's Formula and Selection Bias}.
\newblock \emph{Journal of the American Statistical Association}, 106\penalty0
  (496):\penalty0 1602--1614, 2011.

\bibitem[Epstein et~al.(2023)Epstein, Jabri, Poole, Efros, and
  Holynski]{selfguidance}
Dave Epstein, Allan Jabri, Ben Poole, Alexei~A. Efros, and Aleksander Holynski.
\newblock {Diffusion Self-Guidance for Controllable Image Generation}.
\newblock In \emph{Advances in Neural Information Processing Systems
  (NeurIPS)}, pages 16222--16239, 2023.

\bibitem[Esser et~al.(2024)Esser, Kulal, Blattmann, Entezari, M{\"u}ller,
  Saini, Levi, Lorenz, Sauer, Boesel, Podell, Dockhorn, English, and
  Rombach]{sd3}
Patrick Esser, Sumith Kulal, Andreas Blattmann, Rahim Entezari, Jonas
  M{\"u}ller, Harry Saini, Yam Levi, Dominik Lorenz, Axel Sauer, Frederic
  Boesel, Dustin Podell, Tim Dockhorn, Zion English, and Robin Rombach.
\newblock {Scaling Rectified Flow Transformers for High-Resolution Image
  Synthesis}.
\newblock In \emph{Proceedings of the 41st International Conference on Machine
  Learning (ICML)}, pages 12606--12633, 2024.

\bibitem[Eyring et~al.(2024)Eyring, Karthik, Roth, Dosovitskiy, and
  Akata]{reno}
Luca Eyring, Shyamgopal Karthik, Karsten Roth, Alexey Dosovitskiy, and Zeynep
  Akata.
\newblock {ReNO: Enhancing One-step Text-to-Image Models through Reward-based
  Noise Optimization}.
\newblock In \emph{Advances in Neural Information Processing Systems
  (NeurIPS)}, pages 125487--125519, 2024.

\bibitem[Gao et~al.(2025)Gao, Zha, Zhang, Xue, and Boning]{reg}
Zhengqi Gao, Kaiwen Zha, Tianyuan Zhang, Zihui Xue, and Duane~S. Boning.
\newblock {REG: Rectified Gradient Guidance for Conditional Diffusion Models}.
\newblock In \emph{International Conference on Machine Learning (ICML)}, pages
  18724--18743, 2025.

\bibitem[Go et~al.(2026)Go, Chung, Truong, Bhat, Mi, An, Zhao, Narnhofer,
  Belongie, Tombari, and Schindler]{stitchvm}
Hyojun Go, Hyungjin Chung, Prune Truong, Goutam Bhat, Li Mi, Zhaochong An,
  Zixiang Zhao, Dominik Narnhofer, Serge Belongie, Federico Tombari, and Konrad
  Schindler.
\newblock {Stitched Value Model for Diffusion Alignment}.
\newblock In \emph{Advances in Neural Information Processing Systems
  (NeurIPS)}, 2026.

\bibitem[Hessel et~al.(2021)Hessel, Holtzman, Forbes, Bras, and
  Choi]{clipscore}
Jack Hessel, Ari Holtzman, Maxwell Forbes, Ronan~Le Bras, and Yejin Choi.
\newblock {CLIPScore: A Reference-free Evaluation Metric for Image Captioning}.
\newblock In \emph{Proceedings of the Conference on Empirical Methods in
  Natural Language Processing (EMNLP)}, pages 7514--7528, 2021.

\bibitem[Ho and Salimans(2022)]{cfg}
Jonathan Ho and Tim Salimans.
\newblock {Classifier-Free Diffusion Guidance}.
\newblock arXiv preprint arXiv:2207.12598, 2022.

\bibitem[Ho et~al.(2020)Ho, Jain, and Abbeel]{ddpm}
Jonathan Ho, Ajay Jain, and Pieter Abbeel.
\newblock {Denoising Diffusion Probabilistic Models}.
\newblock In \emph{Advances in Neural Information Processing Systems
  (NeurIPS)}, pages 6840--6851, 2020.

\bibitem[Hong(2024)]{seg}
Susung Hong.
\newblock {Smoothed Energy Guidance: Guiding Diffusion Models with Reduced
  Energy Curvature of Attention}.
\newblock In \emph{Advances in Neural Information Processing Systems
  (NeurIPS)}, pages 66743--66772, 2024.

\bibitem[Hong et~al.(2023)Hong, Lee, Jang, and Kim]{sag}
Susung Hong, Gyuseong Lee, Wooseok Jang, and Seungryong Kim.
\newblock {Improving Sample Quality of Diffusion Models Using Self-Attention
  Guidance}.
\newblock In \emph{Proceedings of the IEEE/CVF International Conference on
  Computer Vision (ICCV)}, pages 7462--7471, 2023.

\bibitem[Huang et~al.(2024)Huang, He, Yu, Zhang, Si, Jiang, Zhang, Wu, Jin,
  Chanpaisit, Wang, Chen, Wang, Lin, Qiao, and Liu]{vbench}
Ziqi Huang, Yinan He, Jiashuo Yu, Fan Zhang, Chenyang Si, Yuming Jiang, Yuanhan
  Zhang, Tianxing Wu, Qingyang Jin, Nattapol Chanpaisit, Yaohui Wang, Xinyuan
  Chen, Limin Wang, Dahua Lin, Yu Qiao, and Ziwei Liu.
\newblock {VBench: Comprehensive Benchmark Suite for Video Generative Models}.
\newblock In \emph{Proceedings of the IEEE/CVF Conference on Computer Vision
  and Pattern Recognition (CVPR)}, pages 21807--21818, 2024.

\bibitem[Hwang et~al.(2026)Hwang, Min, Kim, Shen, and Sung]{noisetilt}
Jisung Hwang, Yunhong Min, Jaihoon Kim, I-Chao Shen, and Minhyuk Sung.
\newblock {NoiseTilt: Noise-Tilted Reverse Kernels for Diffusion Reward
  Alignment}.
\newblock In \emph{{Computer Vision -- ECCV 2026}}, pages 134--153. Springer,
  2026.

\bibitem[Hyung et~al.(2025)Hyung, Kim, Hong, Kim, and Choo]{stg}
Junha Hyung, Kinam Kim, Susung Hong, Min-Jung Kim, and Jaegul Choo.
\newblock {Spatiotemporal Skip Guidance for Enhanced Video Diffusion Sampling}.
\newblock In \emph{Proceedings of the IEEE/CVF Conference on Computer Vision
  and Pattern Recognition (CVPR)}, pages 11006--11015, 2025.

\bibitem[Jain et~al.(2025)Jain, Sareen, Pedramfar, and Ravanbakhsh]{dts}
Vineet Jain, Kusha Sareen, Mohammad Pedramfar, and Siamak Ravanbakhsh.
\newblock {Diffusion Tree Sampling: Scalable inference-time alignment of
  diffusion models}.
\newblock In \emph{Advances in Neural Information Processing Systems
  (NeurIPS)}, pages 145576--145615, 2025.

\bibitem[Jang et~al.(2026)Jang, Ki, Jo, Yoon, Kim, Lin, and
  Hwang]{frameguidance}
Sangwon Jang, Taekyung Ki, Jaehyeong Jo, Jaehong Yoon, Soo~Ye Kim, Zhe Lin, and
  Sung~Ju Hwang.
\newblock {Frame Guidance: Training-Free Guidance for Frame-Level Control in
  Video Diffusion Models}.
\newblock In \emph{International Conference on Learning Representations
  (ICLR)}, pages 60977--61003, 2026.

\bibitem[Karras et~al.(2024)Karras, Aittala, Kynk{\"a}{\"a}nniemi, Lehtinen,
  Aila, and Laine]{autoguidance}
Tero Karras, Miika Aittala, Tuomas Kynk{\"a}{\"a}nniemi, Jaakko Lehtinen, Timo
  Aila, and Samuli Laine.
\newblock {Guiding a Diffusion Model with a Bad Version of Itself}.
\newblock In \emph{Advances in Neural Information Processing Systems}, pages
  52996--53021, 2024.

\bibitem[Kim et~al.(2026)Kim, Lee, Kim, and de~Veciana]{lags}
Heasung Kim, Taekyun Lee, Hyeji Kim, and Gustavo de Veciana.
\newblock {Efficient Weighted Sampling via Score-based Generative Models}.
\newblock In \emph{Proceedings of the IEEE/CVF Conference on Computer Vision
  and Pattern Recognition (CVPR)}, pages 1155--1166, 2026.

\bibitem[Kim(2026)]{asag}
Kwanyoung Kim.
\newblock {Toward the Frontiers of Reliable Diffusion Sampling via Adversarial
  Sinkhorn Attention Guidance}.
\newblock In \emph{Proceedings of the AAAI Conference on Artificial
  Intelligence}, pages 5682--5690, 2026.

\bibitem[Kim and Sim(2025)]{pladis}
Kwanyoung Kim and Byeongsu Sim.
\newblock {PLADIS: Pushing the Limits of Attention in Diffusion Models at
  Inference Time by Leveraging Sparsity}.
\newblock In \emph{Proceedings of the IEEE/CVF International Conference on
  Computer Vision (ICCV)}, pages 16238--16248, 2025.

\bibitem[Kim and Ye(2021)]{noise2score}
Kwanyoung Kim and Jong~Chul Ye.
\newblock {Noise2Score: Tweedie's Approach to Self-Supervised Image Denoising
  without Clean Images}.
\newblock In \emph{Advances in Neural Information Processing Systems}, pages
  864--874, 2021.

\bibitem[Kim et~al.(2025)Kim, Kim, and Park]{das}
Sunwoo Kim, Minkyu Kim, and Dongmin Park.
\newblock {Test-time Alignment of Diffusion Models without Reward
  Over-optimization}.
\newblock In \emph{International Conference on Learning Representations
  (ICLR)}, pages 87238--87272, 2025.

\bibitem[Kirstain et~al.(2023)Kirstain, Polyak, Singer, Matiana, Penna, and
  Levy]{pickscore}
Yuval Kirstain, Adam Polyak, Uriel Singer, Shahbuland Matiana, Joe Penna, and
  Omer Levy.
\newblock {Pick-a-Pic: An Open Dataset of User Preferences for Text-to-Image
  Generation}.
\newblock In \emph{Advances in Neural Information Processing Systems
  (NeurIPS)}, pages 36652--36663, 2023.

\bibitem[Kong et~al.(2024)Kong, Tian, Zhang, Min, Dai, Zhou, Xiong, Li, Wu,
  Zhang, Wu, Lin, Yuan, Long, Wang, Wang, Li, Huang, Yang, Tan, Wang, Song,
  Bai, Wu, Xue, Wang, Wang, Liu, Li, Li, Wang, Yu, Deng, Li, Chen, Cui, Peng,
  Yu, He, Xu, Zhou, Xu, Tao, Lu, Liu, Zhou, Wang, Yang, Wang, Liu, Jiang, and
  Zhong]{hunyuanvideo}
Weijie Kong, Qi Tian, Zijian Zhang, Rox Min, Zuozhuo Dai, Jin Zhou, Jiangfeng
  Xiong, Xin Li, Bo Wu, Jianwei Zhang, Kathrina Wu, Qin Lin, Junkun Yuan,
  Yanxin Long, Aladdin Wang, Andong Wang, Changlin Li, Duojun Huang, Fang Yang,
  Hao Tan, Hongmei Wang, Jacob Song, Jiawang Bai, Jianbing Wu, Jinbao Xue, Joey
  Wang, Kai Wang, Mengyang Liu, Pengyu Li, Shuai Li, Weiyan Wang, Wenqing Yu,
  Xinchi Deng, Yang Li, Yi Chen, Yutao Cui, Yuanbo Peng, Zhentao Yu, Zhiyu He,
  Zhiyong Xu, Zixiang Zhou, Zunnan Xu, Yangyu Tao, Qinglin Lu, Songtao Liu, Dax
  Zhou, Hongfa Wang, Yong Yang, Di Wang, Yuhong Liu, Jie Jiang, and Caesar
  Zhong.
\newblock {HunyuanVideo: A Systematic Framework For Large Video Generative
  Models}.
\newblock \emph{arXiv preprint arXiv:2412.03603}, 2024.

\bibitem[Konovalova et~al.(2026)Konovalova, Idrisov, and Alanov]{shift}
Nina Konovalova, Ibragim Idrisov, and Aibek Alanov.
\newblock {SHIFT: Steering Hidden Intermediates in Flow Transformers}.
\newblock \emph{arXiv preprint arXiv:2604.09213}, 2026.

\bibitem[Li et~al.(2025)Li, Zhao, Wang, Scalia, Eraslan, Nair, Biancalani, Ji,
  Regev, Levine, and Uehara]{svdd}
Xiner Li, Yulai Zhao, Chenyu Wang, Gabriele Scalia, Gokcen Eraslan, Surag Nair,
  Tommaso Biancalani, Shuiwang Ji, Aviv Regev, Sergey Levine, and Masatoshi
  Uehara.
\newblock {Derivative-Free Guidance in Continuous and Discrete Diffusion Models
  with Soft Value-based Decoding}.
\newblock In \emph{Advances in Neural Information Processing Systems
  (NeurIPS)}, pages 95507--95545, 2025.

\bibitem[Lin et~al.(2014)Lin, Maire, Belongie, Hays, Perona, Ramanan,
  Doll{\'a}r, and Zitnick]{coco}
Tsung-Yi Lin, Michael Maire, Serge Belongie, James Hays, Pietro Perona, Deva
  Ramanan, Piotr Doll{\'a}r, and C.~Lawrence Zitnick.
\newblock {Microsoft COCO: Common Objects in Context}.
\newblock In \emph{European Conference on Computer Vision (ECCV)}, pages
  740--755, 2014.

\bibitem[Lipman et~al.(2023)Lipman, Chen, Ben-Hamu, Nickel, and
  Le]{flowmatching}
Yaron Lipman, Ricky T.~Q. Chen, Heli Ben-Hamu, Maximilian Nickel, and Matthew
  Le.
\newblock {Flow Matching for Generative Modeling}.
\newblock In \emph{International Conference on Learning Representations
  (ICLR)}, 2023.

\bibitem[Liu et~al.(2023)Liu, Gong, and Liu]{rectifiedflow}
Xingchao Liu, Chengyue Gong, and Qiang Liu.
\newblock {Flow Straight and Fast: Learning to Generate and Transfer Data with
  Rectified Flow}.
\newblock In \emph{International Conference on Learning Representations
  (ICLR)}, 2023.

\bibitem[Ma et~al.(2025)Ma, Tong, Jia, Hu, Su, Zhang, Yang, Li, Jaakkola, Jia,
  and Xie]{ma2025scaling}
Nanye Ma, Shangyuan Tong, Haolin Jia, Hexiang Hu, Yu-Chuan Su, Mingda Zhang,
  Xuan Yang, Yandong Li, Tommi Jaakkola, Xuhui Jia, and Saining Xie.
\newblock {Scaling Inference Time Compute for Diffusion Models}.
\newblock In \emph{Proceedings of the IEEE/CVF Conference on Computer Vision
  and Pattern Recognition (CVPR)}, pages 2523--2534, 2025.

\bibitem[Patel et~al.(2025)Patel, Wen, Metaxas, and Yang]{flowchef}
Maitreya Patel, Song Wen, Dimitris~N. Metaxas, and Yezhou Yang.
\newblock {FlowChef: Steering of Rectified Flow Models for Controlled
  Generations}.
\newblock In \emph{Proceedings of the IEEE/CVF International Conference on
  Computer Vision (ICCV)}, pages 15308--15318, 2025.

\bibitem[Podell et~al.(2024)Podell, English, Lacey, Blattmann, Dockhorn,
  M{\"u}ller, Penna, and Rombach]{sdxl}
Dustin Podell, Zion English, Kyle Lacey, Andreas Blattmann, Tim Dockhorn, Jonas
  M{\"u}ller, Joe Penna, and Robin Rombach.
\newblock {SDXL: Improving Latent Diffusion Models for High-Resolution Image
  Synthesis}.
\newblock In \emph{International Conference on Learning Representations
  (ICLR)}, pages 1862--1874, 2024.

\bibitem[Radford et~al.(2021)Radford, Kim, Hallacy, Ramesh, Goh, Agarwal,
  Sastry, Askell, Mishkin, Clark, Krueger, and Sutskever]{clip}
Alec Radford, Jong~Wook Kim, Chris Hallacy, Aditya Ramesh, Gabriel Goh,
  Sandhini Agarwal, Girish Sastry, Amanda Askell, Pamela Mishkin, Jack Clark,
  Gretchen Krueger, and Ilya Sutskever.
\newblock {Learning Transferable Visual Models From Natural Language
  Supervision}.
\newblock In \emph{Proceedings of the 38th International Conference on Machine
  Learning (ICML)}, pages 8748--8763, 2021.

\bibitem[Rajabi et~al.(2025)Rajabi, Mehraban, Sadat, and Taati]{tpg}
Javad Rajabi, Soroush Mehraban, Seyedmorteza Sadat, and Babak Taati.
\newblock {Token Perturbation Guidance for Diffusion Models}.
\newblock In \emph{Advances in Neural Information Processing Systems
  (NeurIPS)}, pages 67153--67175, 2025.

\bibitem[Rombach et~al.(2022)Rombach, Blattmann, Lorenz, Esser, and Ommer]{ldm}
Robin Rombach, Andreas Blattmann, Dominik Lorenz, Patrick Esser, and Bj{\"o}rn
  Ommer.
\newblock {High-Resolution Image Synthesis with Latent Diffusion Models}.
\newblock In \emph{Proceedings of the IEEE/CVF Conference on Computer Vision
  and Pattern Recognition (CVPR)}, pages 10684--10695, 2022.

\bibitem[Schuhmann(2022)]{laionaes}
Christoph Schuhmann.
\newblock {Improved Aesthetic Predictor}.
\newblock
  \url{https://github.com/christophschuhmann/improved-aesthetic-predictor},
  2022.

\bibitem[Singhal et~al.(2025)Singhal, Horvitz, Teehan, Ren, Yu, McKeown, and
  Ranganath]{fk}
Raghav Singhal, Zachary Horvitz, Ryan Teehan, Mengye Ren, Zhou Yu, Kathleen
  McKeown, and Rajesh Ranganath.
\newblock {A General Framework for Inference-time Scaling and Steering of
  Diffusion Models}.
\newblock In \emph{International Conference on Machine Learning (ICML)}, pages
  55810--55827, 2025.

\bibitem[Song et~al.(2023)Song, Zhang, Yin, Mardani, Liu, Kautz, Chen, and
  Vahdat]{lgd}
Jiaming Song, Qinsheng Zhang, Hongxu Yin, Morteza Mardani, Ming-Yu Liu, Jan
  Kautz, Yongxin Chen, and Arash Vahdat.
\newblock {Loss-Guided Diffusion Models for Plug-and-Play Controllable
  Generation}.
\newblock In \emph{International Conference on Machine Learning (ICML)}, pages
  32483--32498, 2023.

\bibitem[Talebi and Milanfar(2018)]{nima}
Hossein Talebi and Peyman Milanfar.
\newblock {NIMA: Neural Image Assessment}.
\newblock \emph{IEEE Transactions on Image Processing}, 27\penalty0
  (8):\penalty0 3998--4011, 2018.

\bibitem[Tang et~al.(2025)Tang, Peng, Tang, Hong, Wang, and Chang]{dno}
Zhiwei Tang, Jiangweizhi Peng, Jiasheng Tang, Mingyi Hong, Fan Wang, and
  Tsung-Hui Chang.
\newblock {Inference-Time Alignment of Diffusion Models with Direct Noise
  Optimization}.
\newblock In \emph{Proceedings of the 42nd International Conference on Machine
  Learning (ICML)}, pages 58905--58930, 2025.

\bibitem[{Team Wan} et~al.(2025){Team Wan}, Wang, Ai, Wen, Mao, Xie, Chen, Yu,
  Zhao, Yang, Zeng, Wang, Zhang, Zhou, Wang, Chen, Zhu, Zhao, Yan, Huang, Feng,
  Zhang, Li, Wu, Chu, Feng, Zhang, Sun, Fang, Wang, Gui, Weng, Shen, Lin, Wang,
  Wang, Zhou, Wang, Shen, Yu, Shi, Huang, Xu, Kou, Lv, Li, Liu, Wang, Zhang,
  Huang, Li, Wu, Liu, Pan, Zheng, Hong, Shi, Feng, Jiang, Han, Wu, and
  Liu]{wan}
{Team Wan}, Ang Wang, Baole Ai, Bin Wen, Chaojie Mao, Chen-Wei Xie, Di Chen,
  Feiwu Yu, Haiming Zhao, Jianxiao Yang, Jianyuan Zeng, Jiayu Wang, Jingfeng
  Zhang, Jingren Zhou, Jinkai Wang, Jixuan Chen, Kai Zhu, Kang Zhao, Keyu Yan,
  Lianghua Huang, Mengyang Feng, Ningyi Zhang, Pandeng Li, Pingyu Wu, Ruihang
  Chu, Ruili Feng, Shiwei Zhang, Siyang Sun, Tao Fang, Tianxing Wang, Tianyi
  Gui, Tingyu Weng, Tong Shen, Wei Lin, Wei Wang, Wei Wang, Wenmeng Zhou, Wente
  Wang, Wenting Shen, Wenyuan Yu, Xianzhong Shi, Xiaoming Huang, Xin Xu, Yan
  Kou, Yangyu Lv, Yifei Li, Yijing Liu, Yiming Wang, Yingya Zhang, Yitong
  Huang, Yong Li, You Wu, Yu Liu, Yulin Pan, Yun Zheng, Yuntao Hong, Yupeng
  Shi, Yutong Feng, Zeyinzi Jiang, Zhen Han, Zhi-Fan Wu, and Ziyu Liu.
\newblock {Wan: Open and Advanced Large-Scale Video Generative Models}.
\newblock arXiv preprint arXiv:2503.20314, 2025.

\bibitem[Wallace et~al.(2023)Wallace, Gokul, Ermon, and Naik]{doodl}
Bram Wallace, Akash Gokul, Stefano Ermon, and Nikhil Naik.
\newblock {End-to-End Diffusion Latent Optimization Improves Classifier
  Guidance}.
\newblock In \emph{Proceedings of the IEEE/CVF International Conference on
  Computer Vision (ICCV)}, pages 7280--7290, 2023.

\bibitem[Wallace et~al.(2024)Wallace, Dang, Rafailov, Zhou, Lou, Purushwalkam,
  Ermon, Xiong, Joty, and Naik]{diffusiondpo}
Bram Wallace, Meihua Dang, Rafael Rafailov, Linqi Zhou, Aaron Lou, Senthil
  Purushwalkam, Stefano Ermon, Caiming Xiong, Shafiq Joty, and Nikhil Naik.
\newblock {Diffusion Model Alignment Using Direct Preference Optimization}.
\newblock In \emph{Proceedings of the IEEE/CVF Conference on Computer Vision
  and Pattern Recognition (CVPR)}, pages 8228--8238, 2024.

\bibitem[Wang et~al.(2024)Wang, He, Li, Li, Yu, Ma, Li, Chen, Chen, Wang, Luo,
  Liu, Wang, Wang, and Qiao]{internvid}
Yi Wang, Yinan He, Yizhuo Li, Kunchang Li, Jiashuo Yu, Xin Ma, Xinhao Li, Guo
  Chen, Xinyuan Chen, Yaohui Wang, Ping Luo, Ziwei Liu, Yali Wang, Limin Wang,
  and Yu Qiao.
\newblock {InternVid: A Large-scale Video-Text Dataset for Multimodal
  Understanding and Generation}.
\newblock In \emph{International Conference on Learning Representations
  (ICLR)}, pages 42055--42079, 2024.

\bibitem[Wu et~al.(2024)Wu, Zhang, Zhang, Chen, Liao, Li, Gao, Wang, Zhang,
  Sun, Yan, Min, Zhai, and Lin]{qalign}
Haoning Wu, Zicheng Zhang, Weixia Zhang, Chaofeng Chen, Liang Liao, Chunyi Li,
  Yixuan Gao, Annan Wang, Erli Zhang, Wenxiu Sun, Qiong Yan, Xiongkuo Min,
  Guangtao Zhai, and Weisi Lin.
\newblock {Q-Align: Teaching LMMs for Visual Scoring via Discrete Text-Defined
  Levels}.
\newblock In \emph{International Conference on Machine Learning (ICML)}, pages
  54015--54029, 2024.

\bibitem[Wu et~al.(2023)Wu, Hao, Sun, Chen, Zhu, Zhao, and Li]{hpsv2}
Xiaoshi Wu, Yiming Hao, Keqiang Sun, Yixiong Chen, Feng Zhu, Rui Zhao, and
  Hongsheng Li.
\newblock {Human Preference Score v2: A Solid Benchmark for Evaluating Human
  Preferences of Text-to-Image Synthesis}.
\newblock arXiv preprint arXiv:2306.09341, 2023.

\bibitem[Xu et~al.(2023)Xu, Liu, Wu, Tong, Li, Ding, Tang, and
  Dong]{imagereward}
Jiazheng Xu, Xiao Liu, Yuchen Wu, Yuxuan Tong, Qinkai Li, Ming Ding, Jie Tang,
  and Yuxiao Dong.
\newblock {ImageReward: Learning and Evaluating Human Preferences for
  Text-to-Image Generation}.
\newblock In \emph{Advances in Neural Information Processing Systems
  (NeurIPS)}, pages 15903--15935, 2023.

\bibitem[Ye et~al.(2024)Ye, Lin, Han, Xu, Liu, Liang, Ma, Zou, and Ermon]{tfg}
Haotian Ye, Haowei Lin, Jiaqi Han, Minkai Xu, Sheng Liu, Yitao Liang, Jianzhu
  Ma, James Zou, and Stefano Ermon.
\newblock {TFG: Unified Training-Free Guidance for Diffusion Models}.
\newblock In \emph{Advances in Neural Information Processing Systems
  (NeurIPS)}, pages 22370--22417, 2024.

\bibitem[Yu et~al.(2023)Yu, Wang, Zhao, Ghanem, and Zhang]{freedom}
Jiwen Yu, Yinhuai Wang, Chen Zhao, Bernard Ghanem, and Jian Zhang.
\newblock {FreeDoM: Training-Free Energy-Guided Conditional Diffusion Model}.
\newblock In \emph{Proceedings of the IEEE/CVF International Conference on
  Computer Vision (ICCV)}, pages 23174--23184, 2023.

\bibitem[Zhang et~al.(2018)Zhang, Isola, Efros, Shechtman, and Wang]{lpips}
Richard Zhang, Phillip Isola, Alexei~A. Efros, Eli Shechtman, and Oliver Wang.
\newblock {The Unreasonable Effectiveness of Deep Features as a Perceptual
  Metric}.
\newblock In \emph{Proceedings of the IEEE Conference on Computer Vision and
  Pattern Recognition (CVPR)}, pages 586--595, 2018.

\bibitem[Zhao et~al.(2023)Zhao, Bai, Rao, Zhou, and Lu]{unipc}
Wenliang Zhao, Lujia Bai, Yongming Rao, Jie Zhou, and Jiwen Lu.
\newblock {UniPC: A Unified Predictor-Corrector Framework for Fast Sampling of
  Diffusion Models}.
\newblock In \emph{Advances in Neural Information Processing Systems
  (NeurIPS)}, pages 49842--49869, 2023.

\bibitem[Zhou et~al.(2026)Zhou, Peng, Xu, Li, Springenberg, Frans, and
  Levine]{qgf}
Zhiyuan Zhou, Andy Peng, Charles Xu, Qiyang Li, Tobias Springenberg, Kevin
  Frans, and Sergey Levine.
\newblock {Test-Time Gradient Guidance of Flow Policies in Reinforcement
  Learning}.
\newblock In \emph{Advances in Neural Information Processing Systems
  (NeurIPS)}, 2026.

\end{thebibliography}
}

\clearpage
\onecolumn
\begin{center}
{\Large\bfseries \thetitle\par}
\vspace{0.4em}
{\large Supplementary Material\par}
\end{center}
\vspace{0.6em}

\appendix
\startcontents[appendix]
\noindent{\large\bfseries Contents}
\printcontents[appendix]{}{1}{}
\vspace{1.2em}

\section{Related Work}
\label{app:related}

\paragraph{Reward fine-tuning.}
Fine-tuning adapts the generator to a reward by reinforcement learning~\cite{ddpo}, reward backpropagation~\cite{draft,imagereward}, preference optimization~\cite{diffusiondpo}, or stochastic optimal control~\cite{adjointmatching}.
RadAG keeps the generator frozen and acts only at inference.

\paragraph{Gradient-based guidance.}
DPS~\cite{dps} differentiates a measurement likelihood evaluated at the clean estimate with respect to the noisy sample, and reward alignment replaces this likelihood with a reward.
Universal guidance~\cite{universal} adds an optimization of the clean estimate and per-step self-recurrence, LGD~\cite{lgd} reduces the bias of the point-estimate guidance with Monte Carlo samples, FreeDoM~\cite{freedom} repeats guided steps within a range of noise levels, and Training-Free Guidance (TFG)~\cite{tfg} unifies such methods in one hyperparameter space.
NoiseTilt~\cite{noisetilt} injects the whitened gradient through the noise of each step, and REG~\cite{reg} rescales the guidance gradient element-wise with a diagonal approximation of the denoiser Jacobian, at the cost of an additional backward pass.
\rev{These methods backpropagate through the entire generator at every step, whereas RadAG stops the backward pass at the attention logits of one block.}

\paragraph{Noise optimization.}
DOODL~\cite{doodl} and D-Flow~\cite{dflow} optimize the initial noise by backpropagating the reward through the whole sampler, and ReNO~\cite{reno} and Direct Noise Optimization (DNO)~\cite{dno} optimize the initial or injected noises against preference rewards.
\rev{Each optimization step costs a full sampling pass and its backward pass, while RadAG modifies the prediction at each step of a single sampling pass.}

\paragraph{Approximate backward passes.}
\rev{Some methods keep the gradient with respect to the noisy sample but avoid the backward pass through the generator, at the cost of approximating the Jacobian of the clean estimate.}
\rev{QGF~\cite{qgf}, for flow policies in reinforcement learning, and FlowChef~\cite{flowchef}, for rectified flow models, evaluate the gradient at the clean estimate and replace this Jacobian with the identity, so neither backpropagates through the generator.}
\rev{LAGS~\cite{lags} instead approximates the product of this Jacobian with the gradient, which involves the Hessian of the log density, by a finite difference of two score evaluations, at the cost of an extra forward pass per step.}
\rev{These approximations discard or estimate how the clean estimate depends on the sample through the generator, whereas RadAG takes the exact gradient at the attention logits of one block with a short backward pass, and uses it to construct the reference rather than to update the sample.}

\paragraph{Learned value models.}
Classifier guidance~\cite{cg} trains a classifier on noisy images, and StitchVM~\cite{stitchvm} stitches the early blocks of a frozen diffusion backbone to a truncated reward model, giving a value model on \rev{noisy samples} that makes gradient guidance cheaper.
These methods train a model for each objective, while RadAG uses the reward model as it is.

\paragraph{Candidate-based search.}
Candidate-based search methods sample several candidates and favor those with high reward, by resampling with soft values at each step~\cite{svdd}, sequential Monte Carlo~\cite{das}, Feynman--Kac particle resampling~\cite{fk}, tree search~\cite{dts}, or search over initial noises~\cite{ma2025scaling}.
Each candidate adds a sampling pass, while RadAG guides a single sample.

\paragraph{\rev{Extrapolation guidance.}}
\rev{Extrapolation guidance moves the prediction away from a reference prediction: CFG~\cite{cfg} takes the reference from an unconditional prediction, and autoguidance~\cite{autoguidance} from a separately trained weaker model.}
\mbox{}Perturbation-based guidance builds this reference from the generator itself, by replacing self-attention maps with the identity~\cite{pag}, blurring the queries~\cite{seg}, blurring the regions that self-attention selects~\cite{sag}, skipping layers~\cite{stg}, or shuffling tokens~\cite{tpg}; PLADIS~\cite{pladis} instead extrapolates between dense and sparse cross-attention within a single pass.
\rev{HeadHunter~\cite{headhunter} greedily selects the attention heads to perturb by scoring} samples generated with each candidate head under a given objective, and other techniques refine the extrapolation itself~\cite{nag,cfgpp}.
\rev{These references follow heuristic rules or are selected before sampling, while RadAG constructs a reward-adversarial reference from the reward gradient at every step.}

\paragraph{Attention-based guidance and activation steering.}
Attend-and-Excite~\cite{attendexcite} and self-guidance~\cite{selfguidance} define losses on attention maps and internal representations, and update the latent with their gradient through the generator.
Activation steering adds fixed directions, computed from contrasting prompts, to intermediate activations, as SHIFT~\cite{shift} does for concept removal in diffusion transformers.
RadAG also acts on attention, but uses the reward gradient at the attention logits to construct a reference instead of updating the latent or adding a fixed direction.

\section{Proofs}
\label{app:proof}
\rev{Throughout, $\boldsymbol{z}_t$, $t$, $c$, and the block $\ell$ are fixed, and $\langle \cdot, \cdot \rangle$ and $\|\cdot\|$ are the inner product and norm summed over all heads, queries, and keys.}
Restated results keep their numbers from the main text.

\paragraph{Normalization.}
Let $\mathcal{K}$ be the set of image keys.
For a logit tensor $\mathbf{X}$, let $\mathbf{X}_r$ be its restriction to region $r \in \{\mathrm{img}, \mathrm{txt}\}$, the logits from the image or text queries to $\mathcal{K}$ over all heads (App.~\ref{app:logit_regions}).
Let $P$ remove the mean over $\mathcal{K}$ in each row,
\begin{equation}
(P\mathbf{X}_r)_{h,i,j} = (\mathbf{X}_r)_{h,i,j} - \frac{1}{|\mathcal{K}|} \sum_{j' \in \mathcal{K}} (\mathbf{X}_r)_{h,i,j'}, \qquad j \in \mathcal{K}.
\label{eq:centering}
\end{equation}
Then
\begin{equation}
\revcolor (\tilde{\boldsymbol{g}}_{S_\ell})_r = \frac{P\, (\boldsymbol{g}_{S_\ell})_r}{\rho_r}, \qquad \rho_r = \mathrm{RMS}\big((\boldsymbol{g}_{S_\ell})_r\big) + \zeta,
\label{eq:normalization}
\end{equation}
\rev{with a small constant $\zeta > 0$ for numerical stability, which Eq.~\eqref{eq:normalize} omits, and $\tilde{\boldsymbol{g}}_{S_\ell}$ is zero on the logits to text keys.}
\rev{Since $P$ only removes a component, the RMS of the logit change $\alpha(\tilde{\boldsymbol{g}}_{S_\ell})_r$ is at most $\alpha$; dividing by $\rho_r$ makes this bound invariant to the scale of the reward, and normalizing the two regions separately applies the same bound to image and text queries.}
\rev{RadAG computes $\tilde{\boldsymbol{g}}_{S_\ell}$ once from the base pass and holds it fixed, so $\mathbf{S}_{\mathrm{adv}}$ moves the logits along a fixed direction scaled by $\alpha$.}

\subsection{Proof of Proposition~\ref{prop:ascent}}
\label{app:proof_ascent}
\noindent\textbf{Proposition~\ref{prop:ascent}} (restated)\textbf{.}
\textit{Assume that $\mathbf{S}\mapsto\boldsymbol{v}_\theta(\boldsymbol{z}_t,t,c;\mathbf{S})$ is twice continuously differentiable near $\mathbf{S}_\ell$ and that $R_t$ is twice continuously differentiable near $\boldsymbol{v}_\theta$.
For fixed $w > 0$, as $\alpha \to 0$,}
\begin{equation*}
R_t(\boldsymbol{v}_{\mathrm{RadAG}}) - R_t(\boldsymbol{v}_\theta) = w\,\Delta + O(\alpha^2),
\end{equation*}
\textit{where $\Delta = \alpha\, \langle \boldsymbol{g}_{S_\ell}, \tilde{\boldsymbol{g}}_{S_\ell} \rangle$.}

\begin{proof}
\rev{Let $\boldsymbol{u} = \tilde{\boldsymbol{g}}_{S_\ell}$, let $\boldsymbol{q} = \nabla_{\boldsymbol{v}} R_t(\boldsymbol{v}_\theta)$, and let $\mathbf{J}$ be the Jacobian of $\mathbf{S} \mapsto \boldsymbol{v}_\theta(\boldsymbol{z}_t, t, c; \mathbf{S})$ at $\mathbf{S} = \mathbf{S}_\ell$.}
\rev{By the chain rule applied to Eq.~\eqref{eq:logit_grad}, $\boldsymbol{g}_{S_\ell} = \mathbf{J}^\top \boldsymbol{q}$.}
\rev{A Taylor expansion of $\boldsymbol{v}_\theta$ at $\mathbf{S} = \mathbf{S}_\ell$ in the direction $-\alpha \boldsymbol{u}$ gives}
\begin{equation}
\revcolor \boldsymbol{v}_{\mathrm{adv}} = \boldsymbol{v}_\theta - \alpha \mathbf{J} \boldsymbol{u} + O(\alpha^2),
\end{equation}
and substituting it into Eq.~\eqref{eq:radag_update} gives
\begin{equation}
\revcolor \boldsymbol{v}_{\mathrm{RadAG}} = \boldsymbol{v}_\theta + w\alpha \mathbf{J} \boldsymbol{u} + O(\alpha^2).
\end{equation}
\rev{A Taylor expansion of $R_t$ at $\boldsymbol{v}_\theta$ then gives}
\begin{equation}
\revcolor R_t(\boldsymbol{v}_{\mathrm{RadAG}}) - R_t(\boldsymbol{v}_\theta) = w\alpha\, \langle \boldsymbol{q}, \mathbf{J} \boldsymbol{u} \rangle + O(\alpha^2) = w\alpha\, \langle \mathbf{J}^\top \boldsymbol{q}, \boldsymbol{u} \rangle + O(\alpha^2) = w\alpha\, \langle \boldsymbol{g}_{S_\ell}, \tilde{\boldsymbol{g}}_{S_\ell} \rangle + O(\alpha^2) = w\,\Delta + O(\alpha^2).
\end{equation}
\rev{In the same way, $R_t(\boldsymbol{v}_{\mathrm{adv}}) - R_t(\boldsymbol{v}_\theta) = -\Delta + O(\alpha^2)$, so the update gains $w$ times the reward that the reference loses.}
\end{proof}

\subsection{Sign of the First-Order Term}
\label{app:proof_sign}
\begin{lemma}
\label{lem:normalization_sign}
\rev{$\langle \boldsymbol{g}_{S_\ell},\tilde{\boldsymbol{g}}_{S_\ell}\rangle\geq0$, with equality if and only if $\tilde{\boldsymbol{g}}_{S_\ell}=\mathbf{0}$.}
\end{lemma}

\begin{proof}
\rev{$\tilde{\boldsymbol{g}}_{S_\ell}$ is zero outside the two regions, so}
\begin{equation}
\revcolor \langle \boldsymbol{g}_{S_\ell}, \tilde{\boldsymbol{g}}_{S_\ell} \rangle = \sum_{r} \frac{\langle (\boldsymbol{g}_{S_\ell})_r, P (\boldsymbol{g}_{S_\ell})_r \rangle}{\rho_r} = \sum_{r} \frac{\| P (\boldsymbol{g}_{S_\ell})_r \|^2}{\rho_r},
\end{equation}
where the second equality holds since $P$ is an orthogonal projection, $P^\top = P$ and $P^2 = P$.
\rev{Each term is nonnegative since $\rho_r > 0$, and the sum is zero if and only if $P (\boldsymbol{g}_{S_\ell})_r = 0$ for both regions, that is, $\tilde{\boldsymbol{g}}_{S_\ell} = \mathbf{0}$.}
\end{proof}
\rev{With Proposition~\ref{prop:ascent}, the first-order change of the reward is therefore positive whenever $\tilde{\boldsymbol{g}}_{S_\ell} \neq \mathbf{0}$.}
\rev{If $\tilde{\boldsymbol{g}}_{S_\ell} = \mathbf{0}$, then $\mathbf{S}_{\mathrm{adv}} = \mathbf{S}_\ell$, $\boldsymbol{v}_{\mathrm{adv}} = \boldsymbol{v}_\theta$, and RadAG leaves the prediction unchanged.}
\rev{The centering $P$ removes the component of $\boldsymbol{g}_{S_\ell}$ that shifts all image-key logits of a query equally.}
Since $\texttt{Softmax}$ runs over text and image keys, such a shift only moves attention between the two groups of keys.

\section{Scope and Limitations}
\label{app:limitations}
\paragraph{First-order guarantee.}
Proposition~\ref{prop:ascent} holds as $\alpha \to 0$, whereas RadAG uses a finite $\alpha$ chosen on tuning captions.
\rev{For a finite $\alpha$, the reward change can deviate from its first-order term, as with the step size of gradient guidance.}

\paragraph{Choice of block.}
\rev{RadAG fixes one block per generator, chosen on tuning captions: a late block for the transformers and the middle block for SDXL (Apps.~\ref{app:impl} and~\ref{app:sd3}).}
\rev{Another block may suit other generators, rewards, or step counts better.}

\paragraph{Differentiable rewards.}
\rev{RadAG requires the gradient of the reward and therefore applies only to differentiable rewards, whereas candidate-based search also accepts black-box rewards.}

\paragraph{Other generators.}
The main experiments use FLUX.2-klein at $512^2$ with four steps.
\rev{On SD3 Medium with 25 steps, FreeDoM attains a higher PickScore than RadAG but takes more than twice as long (App.~\ref{app:sd3}).}
\rev{On SDXL, a UNet, the backward pass from the reward to the middle block still traverses the upsampling path, so RadAG takes about as long as DPS and attains a lower PickScore than FreeDoM, NoiseTilt, and DPS (App.~\ref{app:sd3}).}

\section{Experimental Details}
\label{app:details}

\subsection{Protocol}
\label{app:protocol}
\paragraph{Prompts and seeds.}
We evaluate on the COCO val2014 captions~\cite{coco} with indices 100 to 5099 for Table~\ref{tab:main} and Fig.~\ref{fig:reward_time}, and use the index of each caption as its seed, so all methods share the prompts and initial noises.
\rev{Unless stated otherwise, the other quantitative evaluations use the first 1,000 of these captions.}
Hyperparameters are chosen on the captions with indices 0 to 99, which do not overlap with the evaluation captions.

\paragraph{Sampling.}
\rev{Both FLUX.2-klein generators are distilled for four-step sampling without CFG~\cite{flux2}, so they sample at $512^2$ with four Euler steps and no CFG; SD3 Medium and SDXL use CFG (App.~\ref{app:sd3}), and Wan2.1 is described in App.~\ref{app:video}.}
\rev{The generators run in bfloat16 except SDXL, whose UNet and text encoders run in float16 with the official fp16 weights, and the decoder runs in float32 for all generators.}
\rev{On the image generators, every method that takes a reward gradient uses gradient checkpointing.}
\rev{The reward is computed on the decoded image at the generation resolution.}

\paragraph{Target and held-out rewards.}
PickScore~\cite{pickscore} uses the PickScore\_v1 checkpoint (revision a4e4367) with the processor of CLIP ViT-H-14 trained on LAION-2B (revision 1c2b849) in float16, and the aesthetic reward is the LAION aesthetic predictor~\cite{laionaes} on CLIP ViT-L/14.
For the held-out rewards, we compute CLIP score as 100 times the cosine similarity between the image and prompt embeddings, rather than the rescaled CLIPScore~\cite{clipscore}, with prompts truncated to 77 tokens.
ImageReward~\cite{imagereward} applies an MLP to the output of its BLIP text encoder and normalizes it by the official mean and standard deviation, with prompts truncated to 35 tokens.
HPSv2~\cite{hpsv2} uses the HPS v2.1 checkpoint on ViT-H-14.
Q-Align~\cite{qalign} uses the aesthetic task of the one-align checkpoint, on a scale of 1 to 5, and NIMA~\cite{nima} uses the model trained on AVA, on a scale of 1 to 10.
\rev{Generation uses PyTorch 2.5.1, Transformers 4.46.2, and Diffusers 0.35.2, and PickScore, CLIP score, ImageReward, and HPSv2 are evaluated with PyTorch 2.5.1, Transformers 4.46.2, OpenCLIP 3.3.0, and timm 1.0.29.}
\rev{NIMA runs on pyiqa with PyTorch 2.7.1 and Transformers 5.20.2, and Q-Align runs on the official one-align implementation with PyTorch 2.5.1 and Transformers 4.36.1.}

\paragraph{Time and memory.}
\rev{All experiments run on NVIDIA RTX A6000 GPUs, and every time is per image on one GPU.}
The main results use the 5,000 evaluation captions, and Fig.~\ref{fig:reward_time} reports the time per image of these runs.
Fig.~\ref{fig:cost} and App.~\ref{app:breakdown} measure each stage on the first 1,000 evaluation captions, with a batch of 4 on FLUX.2-klein-4B and 2 on FLUX.2-klein-9B, as in the 5,000-caption runs.
Each stage is timed with CUDA events, and the first two batches on each GPU are excluded as warm-up.
The time includes the final decoding but excludes model loading, text encoding, and saving.
Peak memory is the maximum memory allocated by PyTorch.

\subsection{Concentration Measurement}
\label{app:concentration}
\rev{Fig.~\ref{fig:sensitivity} uses FLUX.2-klein-4B, PickScore, and block $\ell$ of Table~\ref{tab:impl} on the 100 tuning captions (indices 0 to 99), and measures the maps at each of the four steps.}
\rev{We segment the final Vanilla image with Mask2Former~\cite{mask2former} (Swin-L, COCO panoptic) and keep the instances of the 80 COCO thing classes.}
\rev{Each instance mask is average-pooled over the $16\times16$-pixel patches of the $32\times32$ token grid, a token lies on the object if at least half of its patch does, and instances with fewer than four tokens are dropped.}
\rev{Every image token of an instance serves as a query, so each query lies on an object.}

\rev{For each query, the map is a row over the 1,024 image keys.}
\rev{For RadAG, it is the normalized reward sensitivity $|\tilde{\boldsymbol{g}}_{S_\ell}|$ of Eq.~\eqref{eq:normalize}, averaged over heads.}
\rev{For the base model and SEG, it is the attention renormalized over the image keys and averaged over heads; SEG replaces every image query by their mean, so all queries share one row.}
\rev{For PAG, it is the identity row.}
\rev{The key of the query itself is removed from the map and the mask, and ties are broken by a negligible random perturbation.}
\rev{Mass accuracy skips queries whose map sums to zero, which leaves PAG without a value.}
\rev{The object area is the share of the remaining keys that lie inside the object.}
\rev{Each measure, including the object area, is averaged over the queries of an instance, the instances of a prompt, the four steps, and the prompts, so large objects do not dominate.}
\rev{Fig.~\ref{fig:motivation} shows the token nearest the center of the largest instance as the query.}

\subsection{Implementation of RadAG}
\label{app:impl}
\paragraph{Generator split.}
\rev{Table~\ref{tab:impl} lists the structure of both FLUX.2-klein generators and the settings of RadAG.}
\rev{FLUX.2-klein~\cite{flux2} consists of double-stream blocks} followed by single-stream blocks, and we index blocks from 0.
At $512^2$, joint attention runs over 512 text tokens and 1,024 image tokens, and the logits are computed after the normalization and rotary embedding of the queries and keys.
RadAG replaces the attention processor of block $\ell$ only.
We select a late block, leaving only the last 4 of 25 blocks of FLUX.2-klein-4B or the last 5 of 32 blocks of FLUX.2-klein-9B for recomputation.

\paragraph{Logit regions.}
\label{app:logit_regions}
In joint attention~\cite{sd3}, the queries and keys include both text and image tokens.
\rev{We modify the logits from every query to the image keys and treat the rows of image queries and of text queries as two regions for the normalization of Eq.~\eqref{eq:normalize}.}
The logits to text keys are unchanged, and $\texttt{Softmax}$ still runs over all keys.

\begin{table}[!htbp]
\centering
\caption{\textbf{Generator structure and settings of RadAG.} Block indices start at 0.}
\label{tab:impl}
\small
\begin{tabular}{@{}lcc@{}}
\toprule
 & FLUX.2-klein-4B & FLUX.2-klein-9B \\
\cmidrule(r){1-1}\cmidrule(lr){2-2}\cmidrule(l){3-3}
Double-stream + single-stream blocks & 5 + 20 & 8 + 24 \\
Heads $\times$ head dimension & $24 \times 128$ & $32 \times 128$ \\
Block $\ell$ (single-stream index / overall index) & 16 / 21 & 19 / 27 \\
Blocks recomputed for $\boldsymbol{v}_{\mathrm{adv}}$ & 4 of 25 & 5 of 32 \\
Logit regions & image and text queries to image keys & image and text queries to image keys \\
Perturbation strength $\alpha$ & 5 & 10 \\
Guidance scale $w$ & 1 & 1 \\
\bottomrule
\end{tabular}
\end{table}

\paragraph{Straight-through attention.}
\label{app:ste}
\rev{Fused attention kernels~\cite{flashattention} do not materialize the logits that the partial backward pass needs.}
\rev{At block $\ell$, RadAG combines the fused output $\widehat{\mathrm{Attn}}_\ell$, computed without gradients, with the explicit output $\widetilde{\mathrm{Attn}}_\ell$ from the logit leaf in a straight-through manner~\cite{ste}:}
\begin{equation}
\revcolor \widetilde{\mathrm{Attn}}_\ell=\texttt{Softmax}(\mathbf{S}_\ell)\mathbf{V}_\ell,\qquad
\mathrm{Attn}_\ell=\widehat{\mathrm{Attn}}_\ell+\widetilde{\mathrm{Attn}}_\ell-\operatorname{sg}(\widetilde{\mathrm{Attn}}_\ell),
\label{eq:ste}
\end{equation}
\rev{where $\operatorname{sg}$ stops the gradient, so $\mathrm{Attn}_\ell$ takes the value of $\widehat{\mathrm{Attn}}_\ell$ in the forward pass and the gradient of $\widetilde{\mathrm{Attn}}_\ell$ in the backward pass.}
\rev{The reference pass uses $\texttt{Softmax}(\mathbf{S}_{\mathrm{adv}})\mathbf{V}_\ell$ at this block, and all other blocks keep fused attention.}

\paragraph{Logit memory.}
The logit leaf used in Eq.~\eqref{eq:ste} occupies 0.56\,GiB on FLUX.2-klein-4B with a batch of 4 and 0.38\,GiB on FLUX.2-klein-9B with a batch of 2.
\paragraph{Cached inputs.}
The base pass caches the input of block $\ell$: hidden states, modulation parameters, rotary embedding, and timestep embedding.
\rev{Rerunning block $\ell$ from this input recomputes the same $\mathbf{S}_\ell$, so replacing it with $\mathbf{S}_{\mathrm{adv}}$ gives the reference $\boldsymbol{v}_{\mathrm{adv}}$ of Eq.~\eqref{eq:perturbation}.}
\rev{With $\mathbf{S}_{\mathrm{adv}}=\mathbf{S}_\ell$, reuse produces an output bitwise identical to that of a full forward pass.}

\paragraph{Other modification locations.}
\label{app:locations}
\rev{For the values, the attention map, the attention output, and the block output in Table~\ref{tab:location}, a variable $\boldsymbol{y}$ with reward gradient $\boldsymbol{g}$ is modified per sample and region by}
\begin{equation}
\revcolor \boldsymbol{\Delta}=-\alpha_{\mathrm{rel}}\,\mathrm{RMS}(\boldsymbol{y})\,\frac{\boldsymbol{g}}{\max\big(\mathrm{RMS}(\boldsymbol{g}),10^{-12}\big)},
\label{eq:rel_delta}
\end{equation}
\rev{without centering, so its size follows the scale of $\boldsymbol{y}$ at each sample, step, and region.}
The modified attention map is clipped at zero and renormalized over each row of all keys.
Each location uses its own $\alpha_{\mathrm{rel}}$ and $w$, tuned on the tuning captions.

\paragraph{Hyperparameter selection.}
We search the block $\ell$ among the single-stream blocks $\{7, 10, 13, 16\}$ of FLUX.2-klein-4B and $\{8, 12, 16, 19\}$ of FLUX.2-klein-9B, with $\alpha \in \{0.5, 1, 5, 10\}$ and $w \in \{0.5, 1, 5, 10\}$.
\rev{On the tuning captions, we choose the setting with the highest target reward separately for PickScore and the aesthetic reward, and both give the values in Table~\ref{tab:impl}.}

\subsection{RadAG+}
\label{app:radagplus}
\rev{RadAG+ adapts attention-level extrapolation and the separate noise estimate of CFG++ to the base pass and the reward-adversarial reference of RadAG (Sec.~\ref{sec:combine}).}

\paragraph{Attention-output extrapolation.}
\rev{For CFG, NAG~\cite{nag} extrapolates between the attention outputs of a positive and a negative branch rather than between final predictions, normalizes the result, and blends it with the positive output.}
\rev{RadAG+ uses the base pass in place of the positive branch and the reward-adversarial reference in place of the negative branch, so the attention output moves away from the reward-decreasing modification.}
\rev{Let $\mathrm{Attn}_\ell$ and $\mathrm{Attn}_{\mathrm{adv}}$ be the attention outputs at block $\ell$ before the output projection, from the base and reference passes.}
\rev{RadAG+ extrapolates them and blends the normalized result with $\mathrm{Attn}_\ell$:}
\begin{equation}
\begin{aligned}
\widetilde{\mathrm{Attn}}&=\mathrm{Attn}_\ell+\gamma\big(\mathrm{Attn}_\ell-\mathrm{Attn}_{\mathrm{adv}}\big),\\
\nu&=\frac{\|\widetilde{\mathrm{Attn}}\|_1}{\max(\|\mathrm{Attn}_\ell\|_1,\delta)},\\
\mathrm{Attn}_+&=(1-\beta)\mathrm{Attn}_\ell+\beta\min(1,\tau/\nu)\widetilde{\mathrm{Attn}},
\end{aligned}
\label{eq:nag}
\end{equation}
\rev{where $\gamma>0$ controls extrapolation, $\tau>0$ bounds the norm ratio, $\beta\in[0,1]$ controls blending, and $\delta>0$ stabilizes the denominator.}
\rev{Passing $\mathrm{Attn}_+$ through the remaining generator computation gives $\boldsymbol{v}_{\mathrm{attn}}$ in Eq.~\eqref{eq:vattn}, and we write $\boldsymbol{v}_{\mathrm{g}}=\boldsymbol{v}_{\mathrm{RadAG}}+(\boldsymbol{v}_{\mathrm{attn}}-\boldsymbol{v}_\theta)$ for the guided velocity.}

The norms in Eq.~\eqref{eq:nag} sum over heads and channels for each token, and the rescaling factor is 1 when $\nu=0$.
\rev{Before blending, the rescaled extrapolation in Eq.~\eqref{eq:nag} has tokenwise $\ell_1$ norm at most $\tau\max(\|\mathrm{Attn}_\ell\|_1,\delta)$.}
\rev{The final attention output therefore has norm at most $(1-\beta)\|\mathrm{Attn}_\ell\|_1+\beta\tau\max(\|\mathrm{Attn}_\ell\|_1,\delta)$.}
\paragraph{Separate clean and noise estimates.}
\rev{For CFG, CFG++~\cite{cfgpp} uses the guided prediction only to predict the clean sample and the unconditional prediction to predict the noise that the update adds back to the latent.}
\rev{RadAG+ applies this separation to flow matching with the base prediction in place of the unconditional one.}
\rev{RadAG+ uses $\boldsymbol{v}_{\mathrm{g}}$ for the clean estimate and $\boldsymbol{v}_\theta$ for the noise estimate:}
\begin{equation}
\begin{aligned}
\hat{\boldsymbol{z}}_{0|t}&=\boldsymbol{z}_t-t\,\boldsymbol{v}_{\mathrm{g}},\qquad
\hat{\boldsymbol{\epsilon}}=\boldsymbol{z}_t+(1-t)\,\boldsymbol{v}_\theta,\\
\boldsymbol{z}_{t'}&=(1-t')\,\hat{\boldsymbol{z}}_{0|t}+t'\,\hat{\boldsymbol{\epsilon}}.
\end{aligned}
\label{eq:cfgpp}
\end{equation}

Substituting the two estimates into the update gives
\begin{equation}
\begin{aligned}
\boldsymbol{z}_{t'}
&=(1-t')\big(\boldsymbol{z}_t-t\,\boldsymbol{v}_{\mathrm{g}}\big)
+t'\big(\boldsymbol{z}_t+(1-t)\boldsymbol{v}_\theta\big)\\
&=\boldsymbol{z}_t-t(1-t')\boldsymbol{v}_{\mathrm{g}}+t'(1-t)\boldsymbol{v}_\theta\\
&=\boldsymbol{z}_t+(t'-t)\Big[\boldsymbol{v}_\theta+\kappa_t\big(\boldsymbol{v}_{\mathrm{g}}-\boldsymbol{v}_\theta\big)\Big],
\end{aligned}
\end{equation}
where $\kappa_t=t(1-t')/(t-t')$ for $t'<t$.
\rev{Since $\boldsymbol{v}_{\mathrm{g}}-\boldsymbol{v}_\theta=w(\boldsymbol{v}_\theta-\boldsymbol{v}_{\mathrm{adv}})+\boldsymbol{v}_{\mathrm{attn}}-\boldsymbol{v}_\theta$, the bracket equals $\boldsymbol{v}_{\mathrm{RadAG+}}$ in Eq.~\eqref{eq:nag_guided}, and the update needs no additional model evaluation.}

\paragraph{Settings.}
\rev{RadAG+ uses Eqs.~\eqref{eq:vattn} and~\eqref{eq:nag_guided} together with the block, regions, and $\alpha$ of Table~\ref{tab:impl}.}
\rev{It uses $\tau=2.5$ and $\beta=0.25$, the default values of NAG~\cite{nag} for FLUX, and $w=0.5$; in Table~\ref{tab:plus}, the variants without attention-output extrapolation keep $w=1$ of RadAG.}
We choose $\gamma$ on the tuning captions, giving $\gamma=4$ for PickScore and $\gamma=6$ for the aesthetic reward.
On FLUX.2-klein-9B, RadAG+ adds about 7\% to the time per image of RadAG.

\begin{table}[!htbp]\centering
\caption{\rev{\textbf{Components of RadAG+.} Mean scores over 1,000 COCO prompts. Attn.\ and Noise denote the two refinements of Sec.~\ref{sec:combine}. Time in s/image; bold marks the best per block.}}
\label{tab:plus_full}
\footnotesize
\setlength{\tabcolsep}{3pt}\renewcommand{\arraystretch}{0.95}
\begin{tabularx}{\textwidth}{@{}cc*{10}{>{\centering\arraybackslash}X}@{}}
\toprule
& & \multicolumn{5}{c}{\textbf{FLUX.2-klein-4B}} & \multicolumn{5}{c}{\textbf{FLUX.2-klein-9B}}\\
\cmidrule(lr){3-7}\cmidrule(l){8-12}
\multicolumn{12}{@{}l}{\textbf{(a) PickScore guidance}}\\
Attn. & Noise & Pick$\uparrow$ & CLIP$\uparrow$ & IR$\uparrow$ & HPSv2$\uparrow$ & Time$\downarrow$ & Pick$\uparrow$ & CLIP$\uparrow$ & IR$\uparrow$ & HPSv2$\uparrow$ & Time$\downarrow$\\
\cmidrule(r){1-2}\cmidrule(lr){3-7}\cmidrule(l){8-12}
 &  & 25.282 & 26.772 & 1.184 & 0.321 & 1.93 & 25.456 & 26.794 & 1.311 & 0.324 & 2.78\\
 & \ding{51} & 25.420 & \textbf{26.773} & 1.186 & 0.321 & \textbf{1.92} & 25.635 & 26.778 & 1.319 & 0.325 & \textbf{2.77}\\
\ding{51} &  & 25.547 & 26.690 & 1.175 & 0.322 & 2.04 & 25.725 & 26.811 & 1.321 & 0.327 & 2.97\\
\oursrow \ding{51} & \ding{51} & \textbf{25.746} & 26.641 & \textbf{1.188} & \textbf{0.324} & 2.03 & \textbf{25.919} & \textbf{26.870} & \textbf{1.327} & \textbf{0.329} & 2.98\\
\midrule
\multicolumn{12}{@{}l}{\textbf{(b) Aesthetic guidance}}\\
Attn. & Noise & AES v2$\uparrow$ & Q-Align$\uparrow$ & NIMA$\uparrow$ & IR$\uparrow$ & Time$\downarrow$ & AES v2$\uparrow$ & Q-Align$\uparrow$ & NIMA$\uparrow$ & IR$\uparrow$ & Time$\downarrow$\\
\cmidrule(r){1-2}\cmidrule(lr){3-7}\cmidrule(l){8-12}
 &  & 6.501 & 3.097 & 5.316 & 1.091 & \textbf{1.86} & 6.621 & 3.346 & 5.411 & 1.236 & \textbf{2.78}\\
 & \ding{51} & 6.596 & 3.143 & 5.370 & \textbf{1.100} & 1.89 & 6.712 & \textbf{3.390} & 5.453 & 1.240 & \textbf{2.78}\\
\ding{51} &  & 6.608 & 3.155 & 5.371 & 1.068 & 2.00 & 6.701 & 3.347 & 5.436 & \textbf{1.243} & 2.99\\
\oursrow \ding{51} & \ding{51} & \textbf{6.701} & \textbf{3.192} & \textbf{5.407} & 1.079 & 2.00 & \textbf{6.782} & 3.378 & \textbf{5.479} & 1.240 & 2.99\\
\bottomrule
\end{tabularx}
\end{table}

\subsection{Baselines}
\label{app:baselines}
All baselines share the evaluation prompts, initial noise seeds, and numerical precision of RadAG; method-specific sampling choices are listed below.

\paragraph{Hyperparameters.}
\rev{Unless noted below, we start from the settings of the original paper or official implementation of each baseline and tune its hyperparameters on the same tuning captions as RadAG.}
\rev{PAG, SEG, and SAG act on block $\ell$ of RadAG, since transformer generators have no middle block (Sec.~\ref{sec:setup}).}
\rev{On FLUX.2-klein, FreeDoM applies time travel at the third of the four steps.}
\rev{NoiseTilt requires injected noise, so it runs on an Euler--Maruyama discretization of the reverse SDE on the timestep grid of each model's default scheduler.}
\rev{SVDD-PM draws five candidates per step with a stochastic transition and keeps the one with the highest clean-estimate reward, the greedy limit of soft-value resampling.}
\rev{Without CFG, each perturbation-based attention guidance method extrapolates by Eq.~\eqref{eq:extrapolation} with its own perturbed prediction.}

\paragraph{PAG.}
At block $\ell$, the logits from image queries to image keys are reduced to their diagonal, so each image query attends to itself and to all text keys.
The rows of text queries are unchanged.
This adapts the identity map of PAG to joint attention, in which the text keys remain.

\paragraph{SEG.}
At block $\ell$, each image query is replaced by the mean of the image queries over the $32 \times 32$ token grid, which is the limit of an infinitely wide Gaussian blur in SEG.
Keys, values, and the rows of text queries are unchanged.

\paragraph{SAG.}
The mask comes from the attention map of block $\ell$ in the base pass.
We renormalize its rows from image queries to image keys, sum the attention each image key receives over the image queries, average this sum over heads, and mark the keys whose sum exceeds 1, its mean over keys.
The mask is upsampled to the latent by nearest neighbors and blurs the clean estimate inside it with a Gaussian kernel of size 9 and standard deviation 1.
The blurred estimate is renoised with the noise of the base prediction, and a full forward pass of the unmodified generator gives the perturbed prediction.

\subsection{Video Generation}
\label{app:video}
\rev{We generate $480 \times 832$ videos of 81 frames with Wan2.1-T2V-1.3B~\cite{wan}, 50 UniPC steps~\cite{unipc}, and CFG scale 5, with the same prompts and initial noise for all methods.}
\rev{The reward is the negative mean squared error between the first and last frames of the clean estimate, decoded from the corresponding latents only, as in the official implementation of Frame Guidance~\cite{frameguidance}.}
\rev{Frame Guidance uses the loop setting of its official implementation, which takes up to seven gradient iterations per step, except that we lower its step size from the default 3 to 1, since the default darkened the videos in our setting.}
\rev{RadAG guides the same steps with one reward gradient per step and modifies the self-attention logits of the 21st of 30 blocks.}
\rev{As in App.~\ref{app:sd3}, only the conditional branch is differentiated and modified, and the null prediction is held fixed.}
\rev{Times exclude model loading and are averaged over eight prompts on one NVIDIA RTX A6000: 374\,s for Vanilla, 1,911\,s for Frame Guidance, and 561\,s for RadAG.}

\paragraph{Quantitative results.}
\rev{Table~\ref{tab:video} evaluates this setting on 100 VBench prompts~\cite{vbench} with seed 42.}
\rev{Loop MSE and Loop LPIPS~\cite{lpips} compare the first and last frames of each video.}
\rev{CLIP-T averages the cosine similarity between the prompt and 16 uniformly spaced frames under CLIP ViT-L/14~\cite{clip}, and ViCLIP-T is the cosine similarity between the prompt and the video embedding of ViCLIP-L/14~\cite{internvid} computed from 8 uniformly spaced frames.}
\rev{Motion is the mean absolute difference between adjacent frames in $[0,1]$ RGB, averaged over pixels, channels, and the 80 adjacent pairs; it is a simple proxy that also responds to camera motion and lighting changes.}
\rev{RadAG attains the lowest loop error on both measures, below that of Frame Guidance, with 3.4 times less time than Frame Guidance.}
\rev{Since the loop reward pulls the last frame toward the first, both guided methods reduce Motion relative to Vanilla by a similar amount, to about 42\% of its value.}
\rev{Both also slightly lower CLIP-T and ViCLIP-T, with RadAG slightly below Frame Guidance on both.}

\begin{table}[!htbp]
\centering
\caption{\rev{\textbf{Looped video generation on Wan2.1-T2V-1.3B} (100 VBench prompts, seed 42). Time in seconds per video, averaged over the eight prompts above. \enabled{} marks the metric that the reward optimizes, and bold marks the best value for each metric except Motion.}}
\label{tab:video}
\small
\begin{tabular}{@{}lrrrrrr@{}}
\toprule
 & \multicolumn{2}{c}{Loop consistency} & \multicolumn{2}{c}{Text alignment} & & \\
\cmidrule(lr){2-3}\cmidrule(lr){4-5}
Method & \textcolor{targetgreen}{MSE$\downarrow$\,$^{\text{\enabled}}$} & LPIPS$\downarrow$ & CLIP-T$\uparrow$ & ViCLIP-T$\uparrow$ & Motion & Time$\downarrow$ \\
\midrule
Vanilla & 0.0393 & 0.3874 & \textbf{0.2433} & \textbf{0.2254} & 0.0422 & \textbf{374} \\
Frame Guidance~\cite{frameguidance} & 0.0049 & 0.1925 & 0.2403 & 0.2168 & 0.0181 & 1,911 \\
\rowcolor{ourshighlight} RadAG (Ours) & \textbf{0.0033} & \textbf{0.1295} & 0.2388 & 0.2129 & 0.0175 & 561 \\
\bottomrule
\end{tabular}
\end{table}

\paragraph{Time breakdown.}
\rev{Table~\ref{tab:video_breakdown} splits the time per video into stages, measured in a separate run on eight prompts.}
\rev{Frame Guidance spends 60\% of its time on backward passes through the generator, whereas RadAG, whose backward pass stops at the 21st block, spends 12\%.}
\rev{With about 32,760 video tokens, the self-attention logits of Wan2.1 do not fit in memory, so RadAG recovers $\boldsymbol{g}_{S_\ell}$ from the gradient of the attention output in chunks of 512 queries.}
\rev{Since the RMS in Eq.~\eqref{eq:normalize} is taken over all logits, this takes one chunked pass to compute the RMS, counted in Other (7\% of the time), and a second one within the reference pass.}

\begin{table}[!htbp]
\centering
\caption{\rev{\textbf{Share of the time per video by stage} (\%, Wan2.1-T2V-1.3B), measured in a separate run on eight prompts. Ref.: reference pass; D+R: decoder and reward. Other includes the chunked pass that computes the RMS for RadAG.}}
\label{tab:video_breakdown}
\small
\begin{tabular}{@{}lrrrrr@{}}
\toprule
 & & & \multicolumn{2}{c}{Backward} & \\
\cmidrule(lr){4-5}
Method & Forward & Ref. & D+R & Generator & Other \\
\midrule
Vanilla & 94 & -- & -- & -- & 6 \\
Frame Guidance~\cite{frameguidance} & 38 & -- & 0.4 & 60 & 2 \\
\rowcolor{ourshighlight} RadAG (Ours) & 61 & 15 & 0.4 & \textbf{12} & 11 \\
\bottomrule
\end{tabular}
\end{table}

\begin{figure*}[!htbp]
\centering
\includegraphics[width=\linewidth]{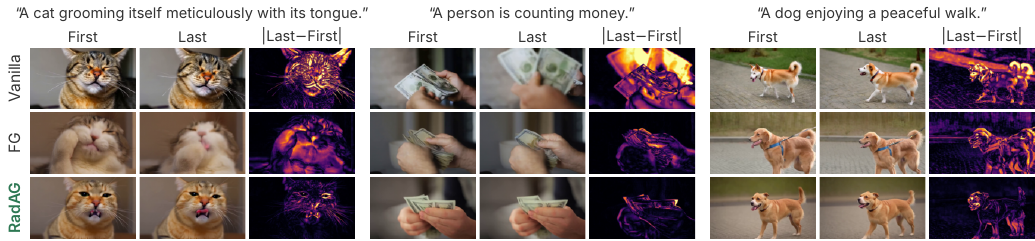}
\caption{\rev{\textbf{Endpoints of looped videos on Wan2.1-T2V-1.3B.} Selected samples from the 100 VBench prompts; $|\mathrm{Last}-\mathrm{First}|$ uses the same color scale as Fig.~\ref{fig:video}, and darker means a more seamless loop. FG: Frame Guidance~\cite{frameguidance}.}}
\label{fig:video_endpoints}
\end{figure*}

\paragraph{Qualitative results.}
\rev{Fig.~\ref{fig:video_endpoints} shows the first and last frames of selected samples.}

\section{Additional Results}
\label{app:results}

\subsection{Other Generators}
\label{app:sd3}
\rev{To show that RadAG applies beyond FLUX.2-klein, we evaluate it on SD3 Medium~\cite{sd3}, a flow-matching transformer, and SDXL~\cite{sdxl}, an $\boldsymbol{\epsilon}$-prediction UNet, under PickScore guidance on 5,000 prompts.}
\rev{Both use CFG with a scale of 7.5 for Vanilla and 5 for guided methods.}
\rev{RadAG modifies only the conditional branch, evaluates the reward on the clean estimate of the CFG prediction, and backpropagates the reward only through the conditional branch, with the null prediction held fixed.}
\rev{The argument of App.~\ref{app:proof} carries over to both generators, since it only uses the differentiability of the reward of the clean estimate in the prediction.}

\paragraph{SD3 Medium.}
\rev{SD3 Medium has the same joint attention as FLUX.2-klein, so RadAG follows Sec.~\ref{sec:method} directly: it modifies the logits from image and text queries to image keys at block 20 of 24 and obtains $\boldsymbol{v}_{\mathrm{adv}}$ by Eq.~\eqref{eq:perturbation}.}
\rev{With the CFG scale $w_{\mathrm{cfg}}$ and the null prediction $\boldsymbol{v}_\varnothing$, the reward is evaluated on the clean estimate of Eq.~\eqref{eq:clean_estimate} with $\boldsymbol{v}_\varnothing+w_{\mathrm{cfg}}(\boldsymbol{v}_\theta-\boldsymbol{v}_\varnothing)$ in place of $\boldsymbol{v}_\theta$, and the guided prediction is}
\begin{equation}
\revcolor \boldsymbol{v}_\varnothing+w_{\mathrm{cfg}}(\boldsymbol{v}_\theta-\boldsymbol{v}_\varnothing)+w(\boldsymbol{v}_\theta-\boldsymbol{v}_{\mathrm{adv}}).
\label{eq:sd3_update}
\end{equation}
\rev{Table~\ref{tab:other_sd3} uses 25 steps at $512^2$ and an empty null prompt.}
RadAG attains the highest CLIP score and takes about 27\% less time than DPS; FreeDoM attains the highest PickScore.

\paragraph{SDXL.}
\rev{SDXL writes the noisy latent as $\boldsymbol{z}_t=\sqrt{\bar\alpha_t}\,\boldsymbol{z}_0+\sqrt{1-\bar\alpha_t}\,\boldsymbol{\epsilon}$ and predicts the noise $\boldsymbol{\epsilon}_\theta$ instead of the velocity.}
\rev{RadAG modifies the logits from image queries to image keys in the self-attention of all ten transformer layers of the middle block, which gave the highest reward in the block search, and normalizes each layer as a separate region in Eq.~\eqref{eq:normalize}.}
\rev{Running the conditional branch with these logits gives the reference $\boldsymbol{\epsilon}_{\mathrm{adv}}$.}
\rev{The reward is evaluated on the clean estimate}
\begin{equation}
\revcolor \hat{\boldsymbol{z}}_{0|t}=\frac{\boldsymbol{z}_t-\sqrt{1-\bar\alpha_t}\,\boldsymbol{\epsilon}_{\mathrm{cfg}}}{\sqrt{\bar\alpha_t}},\qquad
\boldsymbol{\epsilon}_{\mathrm{cfg}}=\boldsymbol{\epsilon}_\varnothing+w_{\mathrm{cfg}}(\boldsymbol{\epsilon}_\theta-\boldsymbol{\epsilon}_\varnothing),
\end{equation}
\rev{which replaces Eq.~\eqref{eq:clean_estimate}, and the guided noise prediction combines the predictions in the same way as Eq.~\eqref{eq:sd3_update},}
\begin{equation}
\revcolor \boldsymbol{\epsilon}_\varnothing+w_{\mathrm{cfg}}(\boldsymbol{\epsilon}_\theta-\boldsymbol{\epsilon}_\varnothing)+w(\boldsymbol{\epsilon}_\theta-\boldsymbol{\epsilon}_{\mathrm{adv}}).
\end{equation}
\rev{Table~\ref{tab:other_sdxl} uses 25 steps at $1024^2$.}
RadAG attains the highest HPSv2, while FreeDoM, NoiseTilt, and DPS attain a higher PickScore.
Its time is close to that of DPS: the backward pass from the reward to the middle block still traverses the entire upsampling path of the UNet, so stopping at the logits saves little.
\rev{On both generators, the reward-based methods share the same batch size, whereas the reward-free methods run with different batch sizes, so the tables report times only for the reward-based methods.}

\paragraph{Baselines.}
\rev{PAG, SEG, and SAG combine the predictions as in Eq.~\eqref{eq:sd3_update} with their own perturbed predictions in place of $\boldsymbol{v}_{\mathrm{adv}}$ or $\boldsymbol{\epsilon}_{\mathrm{adv}}$, and SAG uses the conditional attention map and clean estimate.}
\rev{They act on block 20 of SD3 Medium and the middle block of SDXL, as RadAG does.}
DPS, FreeDoM, and NoiseTilt backpropagate through both branches.

\begin{table}[!htbp]
\centering
\begin{minipage}[t]{0.49\textwidth}
\caption{\textbf{SD3 Medium.} PickScore guidance, \rev{5,000} prompts, \rev{$512^2$, 25 steps}.}
\label{tab:other_sd3}
\centering\fontsize{8.5}{10}\selectfont\setlength{\tabcolsep}{2pt}
\begin{tabularx}{\linewidth}{@{}c@{\hspace{1pt}}c@{\hspace{3pt}}l*{5}{>{\centering\arraybackslash}X}@{}}
& & Method & \textcolor{targetgreen}{Pick$\uparrow$\,$^{\text{\enabled}}$} & CLIP$\uparrow$ & IR$\uparrow$ & HPSv2$\uparrow$ & Time$\downarrow$\\
\cmidrule(r){3-3}\cmidrule(l){4-8}
\multirow{4}{*}{\rotatebox{90}{\scriptsize\color{tablegray} Reward-free}} & \ldelim\{{4}{2.5mm} & Vanilla & 22.051 & 26.298 & 0.863 & 0.278 & --\\
& & SAG~\cite{sag} & 21.911 & 26.327 & 0.820 & 0.278 & --\\
& & PAG~\cite{pag} & 21.826 & 26.513 & 0.766 & 0.272 & --\\
& & SEG~\cite{seg} & 19.910 & 24.949 & $-$0.428 & 0.209 & --\\
\cmidrule(r){3-3}\cmidrule(l){4-8}
\multirow{5}{*}{\rotatebox{90}{\scriptsize\color{tablegray} Reward-based}} & \ldelim\{{5}{2.5mm} & DPS~\cite{dps} & 23.307 & 26.407 & 0.873 & 0.290 & 10.300\\
& & FreeDoM~\cite{freedom} & \textbf{25.591} & 26.531 & \textbf{1.058} & \textbf{0.320} & 17.100\\
& & SVDD-PM~\cite{svdd} & 22.786 & 26.248 & 0.967 & 0.292 & 21.260\\
& & NoiseTilt~\cite{noisetilt} & 24.912 & 26.554 & 0.991 & 0.311 & 12.800\\
& & \cellcolor{ourshighlight}\textbf{RadAG} (Ours) & \cellcolor{ourshighlight}25.266 & \cellcolor{ourshighlight}\textbf{26.875} & \cellcolor{ourshighlight}0.934 & \cellcolor{ourshighlight}0.309 & \cellcolor{ourshighlight}\textbf{7.470}\\
\end{tabularx}
\end{minipage}\hfill
\begin{minipage}[t]{0.49\textwidth}
\caption{\textbf{SDXL.} PickScore guidance, 5,000 prompts, $1024^2$, 25 steps.}
\label{tab:other_sdxl}
\centering\fontsize{8.5}{10}\selectfont\setlength{\tabcolsep}{2pt}
\begin{tabularx}{\linewidth}{@{}c@{\hspace{1pt}}c@{\hspace{3pt}}l*{5}{>{\centering\arraybackslash}X}@{}}
& & Method & \textcolor{targetgreen}{Pick$\uparrow$\,$^{\text{\enabled}}$} & CLIP$\uparrow$ & IR$\uparrow$ & HPSv2$\uparrow$ & Time$\downarrow$\\
\cmidrule(r){3-3}\cmidrule(l){4-8}
\multirow{4}{*}{\rotatebox{90}{\scriptsize\color{tablegray} Reward-free}} & \ldelim\{{4}{2.5mm} & Vanilla & 22.432 & 26.875 & 0.763 & 0.278 & --\\
& & SAG~\cite{sag} & 22.251 & 26.478 & 0.659 & 0.278 & --\\
& & PAG~\cite{pag} & 22.443 & 26.303 & 0.638 & 0.277 & --\\
& & SEG~\cite{seg} & 22.264 & 26.053 & 0.616 & 0.282 & --\\
\cmidrule(r){3-3}\cmidrule(l){4-8}
\multirow{5}{*}{\rotatebox{90}{\scriptsize\color{tablegray} Reward-based}} & \ldelim\{{5}{2.5mm} & DPS~\cite{dps} & 24.687 & 26.687 & 0.784 & 0.297 & 39.971\\
& & FreeDoM~\cite{freedom} & \textbf{25.188} & 26.900 & \textbf{0.886} & 0.309 & 71.646\\
& & SVDD-PM~\cite{svdd} & 23.102 & \textbf{27.134} & 0.828 & 0.283 & 78.675\\
& & NoiseTilt~\cite{noisetilt} & 24.984 & 26.810 & 0.838 & 0.300 & 43.406\\
& & \cellcolor{ourshighlight}\textbf{RadAG} (Ours) & \cellcolor{ourshighlight}24.557 & \cellcolor{ourshighlight}26.332 & \cellcolor{ourshighlight}0.801 & \cellcolor{ourshighlight}\textbf{0.310} & \cellcolor{ourshighlight}\textbf{39.446}\\
\end{tabularx}
\end{minipage}
\par\smallskip
\begin{minipage}{\textwidth}\footnotesize
\enabled marks the target reward; the other scores are held-out rewards. \rev{Bold marks the best guided result.} Time is seconds per image for the reward-based methods.
\end{minipage}
\end{table}

\subsection{Direction of the Modification}
\label{app:direction}
RadAG modifies the logits against the reward gradient to construct a reward-adversarial reference and extrapolates away from it.
\rev{One can instead modify them along the gradient, $\mathbf{S}_\ell+\alpha\tilde{\boldsymbol{g}}_{S_\ell}$, to construct a reward-cooperative reference $\boldsymbol{v}_{\mathrm{coop}}$ and extrapolate toward it as $\boldsymbol{v}_\theta + w\,(\boldsymbol{v}_{\mathrm{coop}} - \boldsymbol{v}_\theta)$.}
By the expansion in the proof of Proposition~\ref{prop:ascent}, both choices change the reward by the same first-order term and differ only in higher-order terms.
Table~\ref{tab:direction} compares them with the same hyperparameters under PickScore guidance on a transformer, FLUX.2-klein-4B, and on a UNet, SDXL~\cite{sdxl}.
On FLUX.2-klein-4B, the reward-cooperative direction attains a slightly higher PickScore and CLIP score, and the reward-adversarial direction attains a higher ImageReward and HPSv2.
On SDXL, the reward-cooperative direction attains lower values than the reward-adversarial direction on all four metrics.
RadAG therefore uses the reward-adversarial direction, \rev{which performs comparably to the reward-cooperative direction on FLUX.2-klein-4B and substantially better on SDXL,} and keeps the extrapolation form of Eq.~\eqref{eq:extrapolation}.

\begin{table}[!htbp]
\centering
\caption{\textbf{Direction of the modification} (PickScore guidance, \rev{1,000 prompts}, same hyperparameters for both directions). Bold marks the better direction for each model. \rev{\enabled marks the target reward; the other scores are held-out rewards.}}
\label{tab:direction}
\small
\begin{tabular}{@{}llrrrr@{}}
\toprule
Model & Direction & \textcolor{targetgreen}{Pick$\uparrow$\,$^{\text{\enabled}}$} & CLIP$\uparrow$ & IR$\uparrow$ & HPSv2$\uparrow$ \\
\cmidrule(r){1-2}\cmidrule(l){3-6}
& Reward-cooperative & \textbf{25.334} & \textbf{26.811} & 1.143 & 0.316 \\
\multirow{-2}{*}{FLUX.2-klein-4B, $512^2$, 4 steps} & \cellcolor{ourshighlight}Reward-adversarial & \cellcolor{ourshighlight}25.282 & \cellcolor{ourshighlight}26.772 & \cellcolor{ourshighlight}\textbf{1.184} & \cellcolor{ourshighlight}\textbf{0.321} \\
\cmidrule(r){1-2}\cmidrule(l){3-6}
& Reward-cooperative & 22.939 & 24.998 & $-$0.052 & 0.258 \\
\multirow{-2}{*}{SDXL, $1024^2$, 25 steps} & \cellcolor{ourshighlight}Reward-adversarial & \cellcolor{ourshighlight}\textbf{24.567} & \cellcolor{ourshighlight}\textbf{26.255} & \cellcolor{ourshighlight}\textbf{0.820} & \cellcolor{ourshighlight}\textbf{0.309} \\
\bottomrule
\end{tabular}
\end{table}

\subsection{Cost Breakdown}
\label{app:breakdown}
\rev{For the reward-based methods in Fig.~\ref{fig:cost}, Table~\ref{tab:breakdown} lists the stage times and the peak memory under PickScore guidance.}

\begin{table}[!htbp]
\centering
\caption{\rev{\textbf{Time per image by stage} (seconds, PickScore guidance, 1,000 prompts, RTX A6000). D+R: decoder and reward model, Ref.: reference pass of block $\ell$ and the later blocks. Other includes the initial setup, the construction of $\mathbf{S}_{\mathrm{adv}}$, the update, and the final decoding. Memory is the peak memory allocated (GiB), with batch size 4 (4B) and 2 (9B). Bold marks the lowest time.}}
\label{tab:breakdown}
\fontsize{8.5}{10}\selectfont
\setlength{\tabcolsep}{4.5pt}
\begin{tabular}{@{}ll*{7}{r}rr@{}}
\toprule
 & & \multicolumn{3}{c}{Forward} & \multicolumn{2}{c}{Backward} & & & & \\
\cmidrule(lr){3-5}\cmidrule(lr){6-7}
Model & Method & Generator & Ref. & D+R & Generator & D+R & Whitening & Other & Total & Memory \\
\cmidrule(r){1-2}\cmidrule(lr){3-5}\cmidrule(lr){6-7}\cmidrule(l){8-11}
\multirow{4}{*}{FLUX.2-klein-4B}
 & DPS~\cite{dps} & 0.520 & -- & 0.500 & 1.133 & 0.480 & -- & 0.120 & 2.753 & 35.140 \\
 & NoiseTilt~\cite{noisetilt} & 0.509 & -- & 0.472 & 1.129 & 0.452 & 0.609 & 0.120 & 3.291 & 35.130 \\
 & SVDD-PM~\cite{svdd} & 2.394 & -- & 2.413 & -- & -- & -- & 0.125 & 4.932 & 13.210 \\
 & \cellcolor{ourshighlight}RadAG (Ours) & \cellcolor{ourshighlight}0.528 & \cellcolor{ourshighlight}0.102 & \cellcolor{ourshighlight}0.496 & \cellcolor{ourshighlight}\textbf{0.186} & \cellcolor{ourshighlight}0.466 & \cellcolor{ourshighlight}-- & \cellcolor{ourshighlight}0.135 & \cellcolor{ourshighlight}\textbf{1.913} & \cellcolor{ourshighlight}35.810 \\
\cmidrule(r){1-2}\cmidrule(lr){3-5}\cmidrule(lr){6-7}\cmidrule(l){8-11}
\multirow{4}{*}{FLUX.2-klein-9B}
 & DPS~\cite{dps} & 1.107 & -- & 0.598 & 2.357 & 0.464 & -- & 0.132 & 4.658 & 32.290 \\
 & NoiseTilt~\cite{noisetilt} & 1.097 & -- & 0.567 & 2.352 & 0.471 & 0.715 & 0.131 & 5.332 & 32.280 \\
 & SVDD-PM~\cite{svdd} & 5.135 & -- & 2.555 & -- & -- & -- & 0.128 & 7.817 & 21.010 \\
 & \cellcolor{ourshighlight}RadAG (Ours) & \cellcolor{ourshighlight}1.125 & \cellcolor{ourshighlight}0.197 & \cellcolor{ourshighlight}0.500 & \cellcolor{ourshighlight}\textbf{0.365} & \cellcolor{ourshighlight}0.465 & \cellcolor{ourshighlight}-- & \cellcolor{ourshighlight}0.149 & \cellcolor{ourshighlight}\textbf{2.801} & \cellcolor{ourshighlight}32.600 \\
\bottomrule
\end{tabular}
\end{table}

\paragraph{Whitening overhead.}
NoiseTilt adds gradient whitening to its backward computation, accounting for 18.5\% and 13.4\% of its total time on FLUX.2-klein-4B and 9B, respectively.
\paragraph{Candidate-based search.}
SVDD-PM needs no backward pass but evaluates five candidates at each step, which takes 19 forward passes of the generator, 21 of the decoder, and 20 of the reward model per image.
\rev{The forward passes of the generator, decoder, and reward model take 98\% of its time on both models.}

\paragraph{Memory.}
SVDD-PM has the lowest peak memory of these methods, as it needs no backward pass.

\subsection{Qualitative Results}
\label{app:qual}
\begin{figure}[p]
\centering
\includegraphics[height=0.86\textheight]{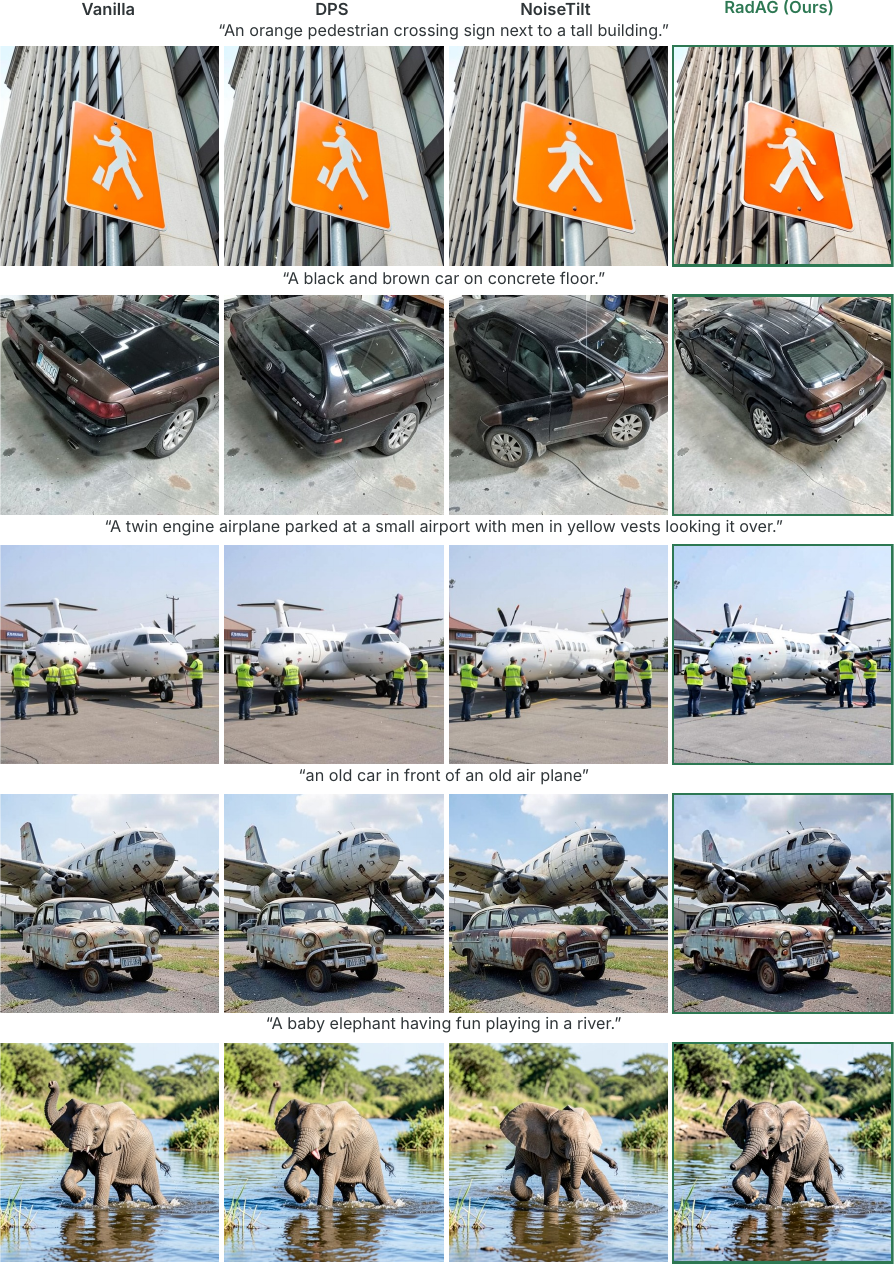}
\caption{\textbf{Qualitative comparison on FLUX.2-klein-4B} ($512^2$, four steps, PickScore guidance). Each row shares the prompt and the seed, and RadAG is outlined in green.}
\label{fig:qual_4b}
\end{figure}

\begin{figure}[p]
\centering
\includegraphics[height=0.86\textheight]{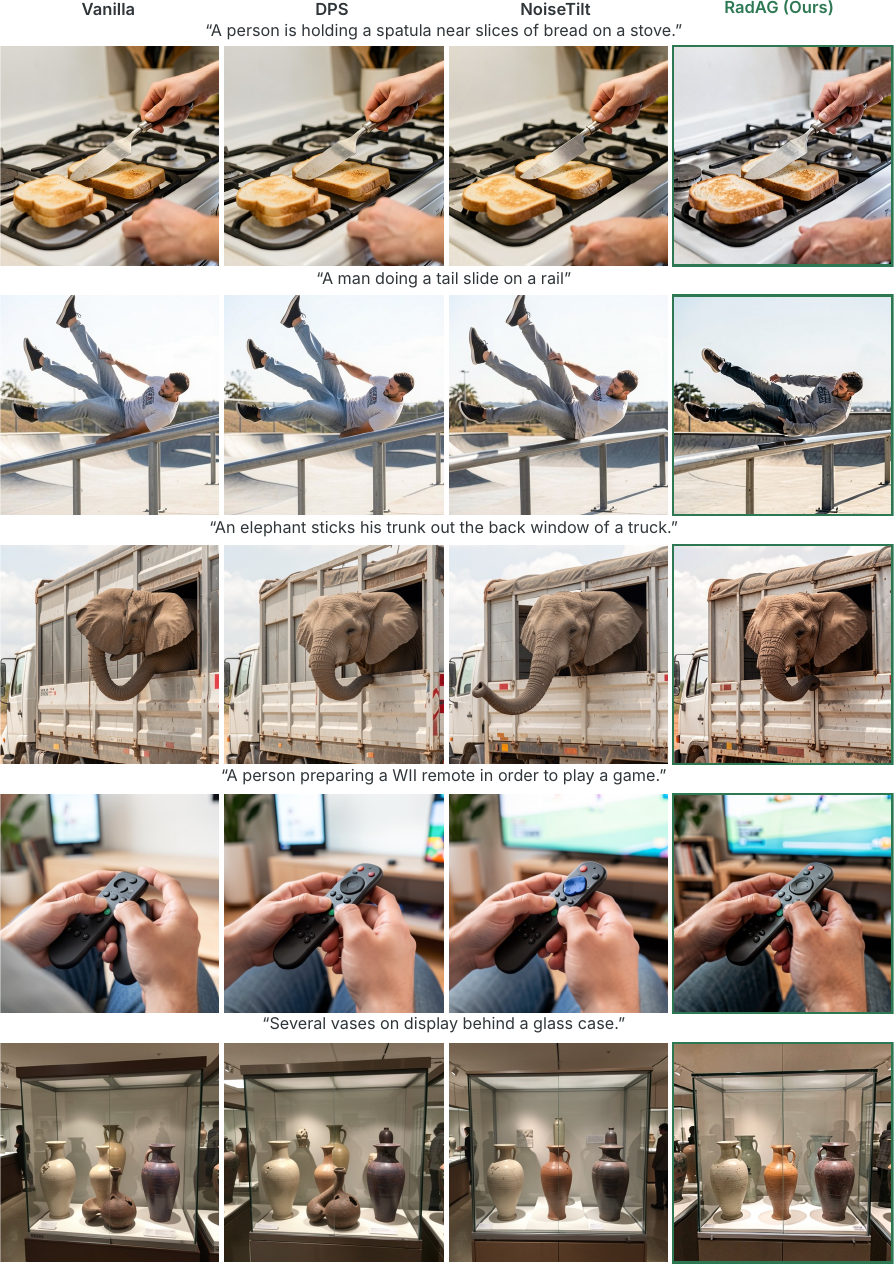}
\caption{\textbf{Qualitative comparison on FLUX.2-klein-9B} ($512^2$, four steps, PickScore guidance). Each row shares the prompt and the seed, and RadAG is outlined in green.}
\label{fig:qual_9b}
\end{figure}

\begin{figure}[p]
\centering
\includegraphics[height=0.86\textheight]{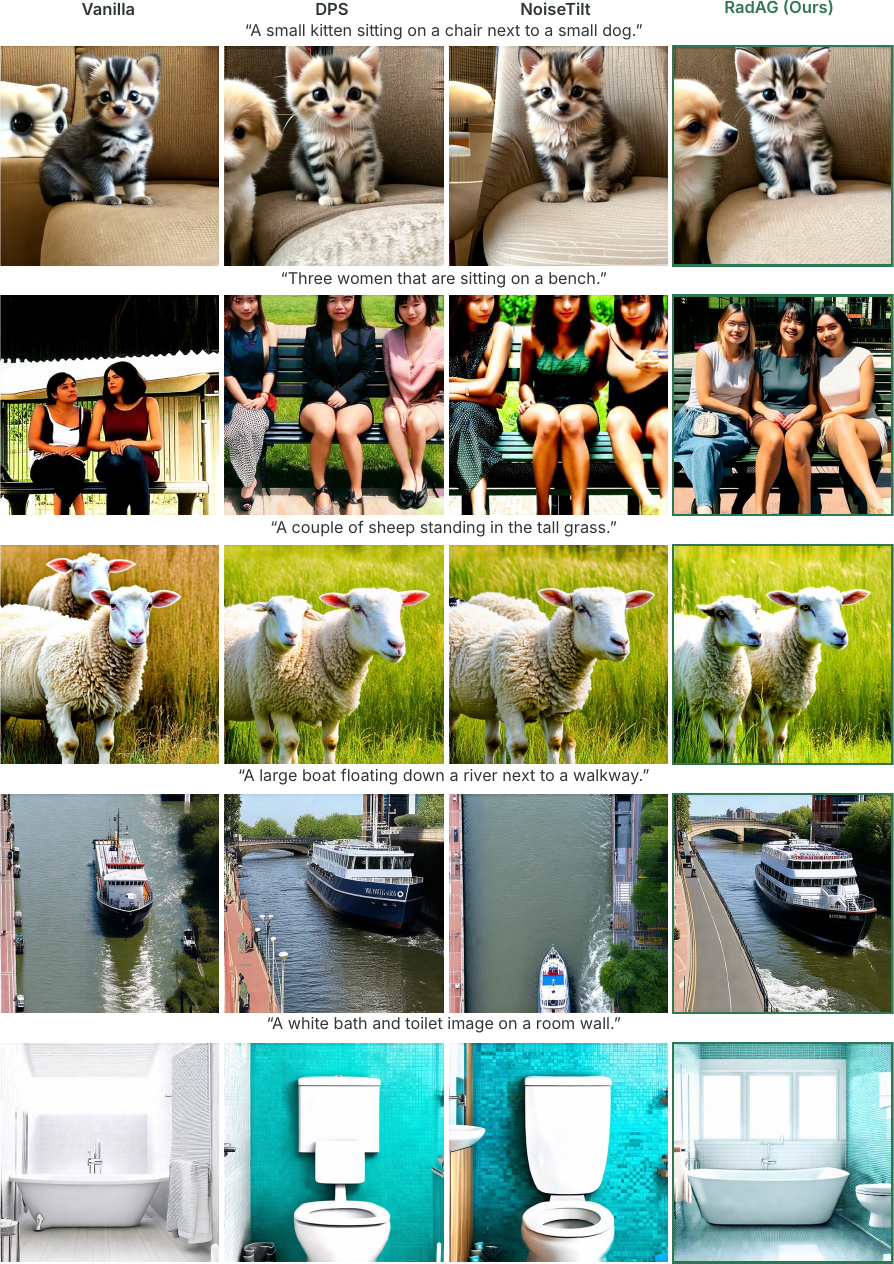}
\caption{\textbf{Qualitative comparison on SD3 Medium} ($512^2$, 25 steps, PickScore guidance). Each row shares the prompt and the seed, and RadAG is outlined in green.}
\label{fig:qual_sd3}
\end{figure}

\begin{figure}[p]
\centering
\includegraphics[height=0.86\textheight]{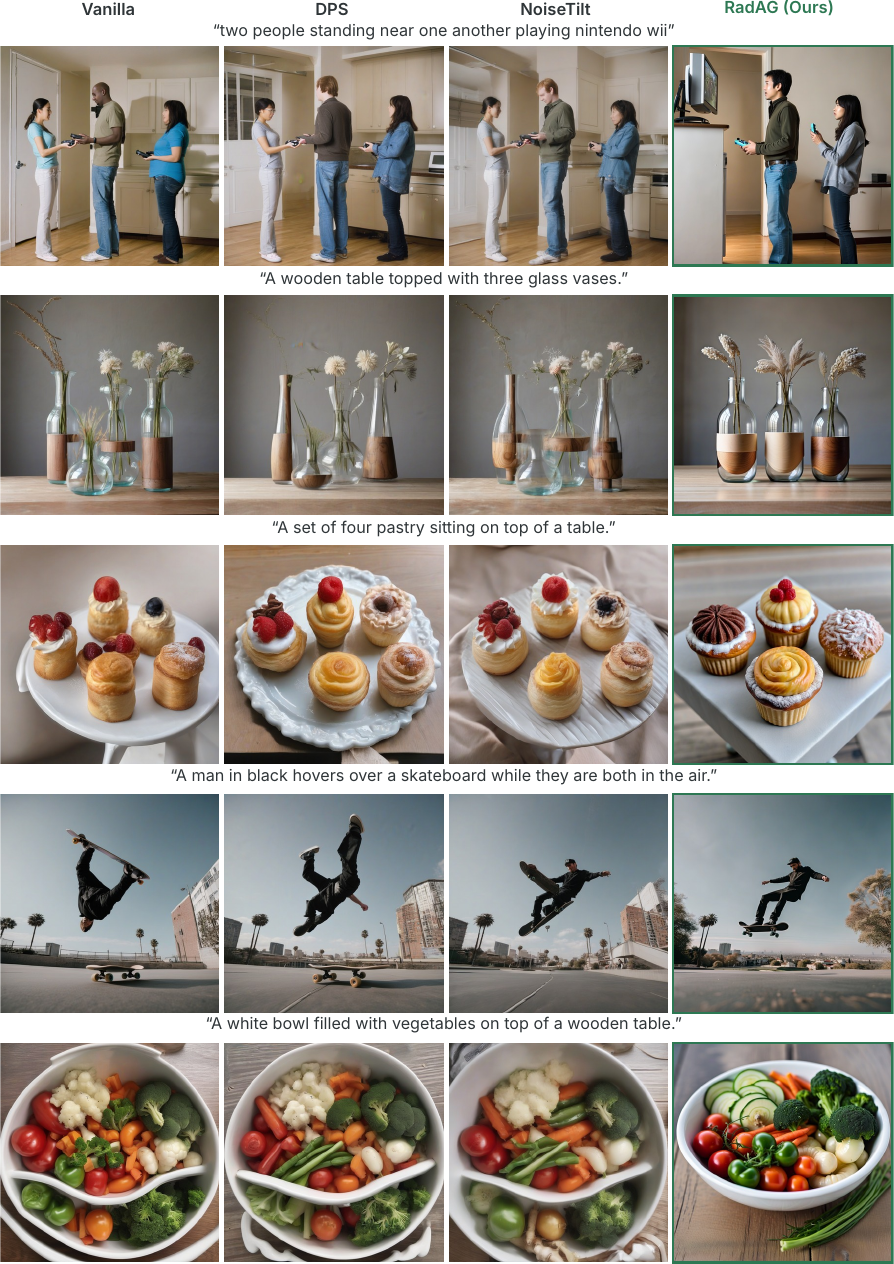}
\caption{\rev{\textbf{Qualitative comparison on SDXL} ($1024^2$, 25 steps, PickScore guidance). Each row shares the prompt and the seed, and RadAG is outlined in green.}}
\label{fig:qual_sdxl}
\end{figure}

Figs.~\ref{fig:qual_4b}, \ref{fig:qual_9b}, \ref{fig:qual_sd3}, and~\ref{fig:qual_sdxl} compare Vanilla, DPS, NoiseTilt, and RadAG on FLUX.2-klein-4B, FLUX.2-klein-9B, SD3 Medium, and \rev{SDXL}.
Each row uses the same prompt and seed for all methods, and the prompts are selected to show visible differences between the methods.

\end{document}